\documentclass[runningheads]{llncs}
\usepackage{eccv}

\usepackage{eccvabbrv}

\usepackage{graphicx}
\usepackage{epsfig}
\usepackage{graphicx}
\usepackage{amsmath}
\usepackage{amssymb}
\usepackage{booktabs}
\usepackage{multirow}
\usepackage{xcolor}
\usepackage{algorithm}
\usepackage{algorithmic}
\usepackage{subcaption}
\usepackage{hyperref}
\usepackage{array}
\usepackage[table]{xcolor}
\usepackage[accsupp]{axessibility}  
\usepackage{amsmath}
\usepackage{amssymb}
\usepackage{makecell}   

\usepackage{hyperref}
\hypersetup{
    colorlinks=true,
    linkcolor=blue,
    filecolor=magenta,
    urlcolor=cyan,
    pdftitle={Your Title},
    pdfpagemode=FullScreen,
}

\usepackage{orcidlink}

\begin{document}
\newcommand{\titleicon}{\raisebox{-0.16\height}{\includegraphics[height=1.15em]{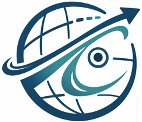}}}
\title{\texorpdfstring{\protect\titleicon}{}WorldCache: Content-Aware Caching for Accelerated Video World Models}

\titlerunning{WorldCache: Content-Aware Caching for Accelerated Video World Models}

\author{Umair Nawaz\inst{1}\thanks{Corresponding author: \email{umair.nawaz@mbzuai.ac.ae}} \and
Ahmed Heakl\inst{1} \and
Ufaq Khan\inst{1} \and Abdelrahman M. Shaker\inst{1} \and \\Salman Khan\inst{1} \and Fahad Shahbaz Khan\inst{1,2}}

\authorrunning{U.~Nawaz et al.}

\institute{Mohamed bin Zayed University of Artificial Intelligence, UAE \and
Link\"{o}ping University, Sweden}

\maketitle

\begin{abstract}
Diffusion Transformers (DiTs) power high-fidelity video world models but remain computationally expensive due to sequential denoising and costly spatio-temporal attention. Training-free feature caching accelerates inference by reusing intermediate activations across denoising steps; however, existing methods largely rely on a Zero-Order Hold assumption i.e., reusing cached features as static snapshots when global drift is small. This often leads to ghosting artifacts, blur, and motion inconsistencies in dynamic scenes.
We propose \textbf{WorldCache}, a Perception-Constrained Dynamical Caching framework that improves both \emph{when} and \emph{how} to reuse features. WorldCache introduces motion-adaptive thresholds, saliency-weighted drift estimation, optimal approximation via blending and warping, and phase-aware threshold scheduling across diffusion steps. Our cohesive approach enables adaptive, motion-consistent feature reuse without retraining.
On Cosmos-Predict2.5-2B evaluated on PAI-Bench, WorldCache achieves \textbf{2.3$\times$} inference speedup while preserving \textbf{99.4\%} of baseline quality, substantially outperforming prior training-free caching approaches. Our project can be accessed on: \href{https://umair1221.github.io/World-Cache/}{World-Cache}.
\end{abstract}

\section{Introduction}
\label{sec:intro}
World models predict future visual states that are physically
consistent and useful for downstream decision-making, enabling
agents to plan and act within simulated
environments~\cite{zhao2025world}. Large-scale Diffusion
Transformers (DiTs) have become the dominant backbone for such
models~\cite{wang2025lavin,yang2024cogvideox,chen2024gentron},
because spatio-temporal attention over latent tokens captures the
long-range dependencies central to world consistency (\eg, object
permanence and causal motion). However, this expressiveness comes
at a steep computational cost: world-model rollouts require many
frames, and each frame is produced by sequentially invoking deep
transformer blocks across dozens of denoising
steps~\cite{ma2025efficient,chi2025mind}. The resulting latency is
the primary obstacle to interactive world simulation and
closed-loop deployment.

A natural remedy is to exploit redundancy along the denoising
trajectory. Consecutive steps often produce only small changes in
intermediate features~\cite{fuest2026diffusion}, so recomputing
every block at every step is wasteful. Training-free caching
methods exploit this observation: they estimate a step-to-step
drift using a lightweight probe, then skip expensive layers when
drift falls below a threshold, reusing cached activations instead.
FasterCache~\cite{lyu2025fastercache} applies this idea to video
DiTs with a fixed skip schedule, and
DiCache~\cite{bu2026dicache} makes it adaptive via shallow-layer
probes that decide both \emph{when} and \emph{how} to reuse cached
states.
\begin{figure}[t]
    \centering
    \includegraphics[width=\linewidth]{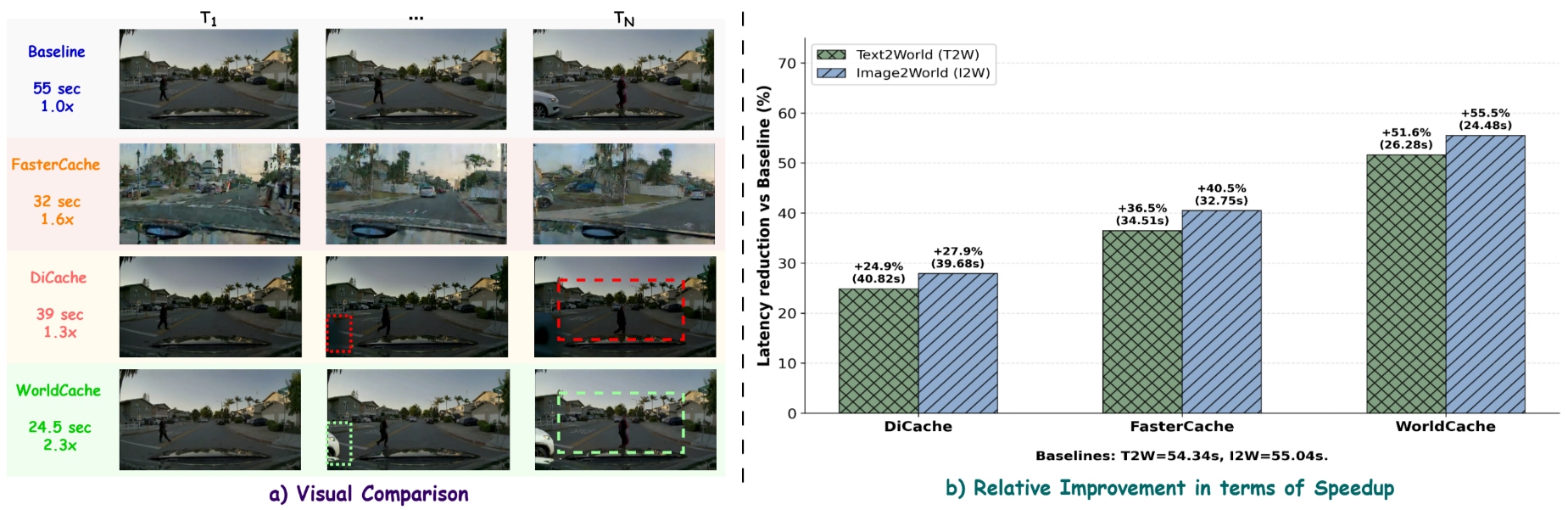}
\caption{\textbf{Qualitative and quantitative comparison of acceleration methods on video world model generation using Cosmos-Predict2.5-2B.} \textbf{Left:} Visual comparison of Baseline (no acceleration), FasterCache, DiCache, and WorldCache (ours) across three representative timesteps ($T_1, \ldots, T_N$) of a driving scene from the City Street domain. FasterCache achieves $1.6\times$ speedup but introduces severe visual artifacts and scene hallucinations. DiCache ($1.3\times$) better preserves scene fidelity but exhibits noticeable spatial artifacts at later timesteps (red dashed boxes). WorldCache (ours) achieves $2.3\times$ speedup while faithfully reproducing scene content, motion, and background structure across all timesteps (green dashed boxes). \textbf{Right:} Relative latency reduction (\%) over Baseline for Text2World (T2W) and Image2World (I2W) tasks, showing WorldCache achieves up to $55.5\%$ reduction compared to $40.5\%$ for FasterCache and $27.9\%$ for DiCache. All results are reported on a single H200 GPU.}
    \label{fig:teaser}
    \vspace{-0.1in}
\end{figure}
For world models, however, this ``skip-and-reuse'' paradigm fails
precisely where it matters most: scenes with significant motion \cite{khan2024deepskinformer}
and salient interactions~\cite{li2025comprehensive}. The failure
has a single root cause. Existing methods treat cache reuse as a
\emph{zero-order hold}: when probe drift is small, they copy stale
features verbatim into the next step. Under motion, this produces
ghosting, semantic smearing, and incoherent trajectories (as shown in Fig.~\ref{fig:teaser}), exactly
the artifacts that break world-model rollouts, where errors
compound across autoregressive generation. Three specific
blindspots make the problem worse. \textit{First}, global drift metrics
average over the entire spatial map, so a static background can
mask large foreground changes, causing the method to skip when it
should recompute. \textit{Second}, all spatial locations are weighted
equally, even though errors on salient entities (agents, hands,
manipulated objects) dominate both perceptual and functional
quality. \textit{Third}, a single static threshold ignores that early
denoising steps establish global structure while late steps only
refine high-frequency detail; a threshold tuned for the early
phase becomes wastefully conservative in the late phase.

We propose \textbf{WorldCache}, a training-free caching framework
that replaces the zero-order hold with a
\emph{perception-constrained dynamical approximation} designed for
DiT-based world models. WorldCache addresses each blindspot above
with a lightweight, composable module. Causal Feature
Caching (CFC) adapts the skip threshold to latent motion
magnitude, preventing stale reuse during fast dynamics.
Saliency-Weighted Drift (SWD) reweights the probe signal
toward perceptually important regions, so caching decisions
reflect foreground fidelity rather than background noise.
Optimal Feature Approximation (OFA) replaces verbatim
copying with least-squares optimal blending and motion-compensated
warping, reducing approximation error when skipping does occur.
Adaptive Threshold Scheduling (ATS) progressively relaxes
the threshold during late denoising, where aggressive reuse is
both safe and highly effective. Together, these modules convert
caching from a brittle shortcut into a controlled approximation
strategy aligned with world-model requirements.

On the Physical AI Bench
(PAI-Bench)~\cite{zhou2025paibench}, WorldCache achieves a
$2.3\times$ speedup on Cosmos-Predict2.5 (2B) while preserving
$99.4\%$ of baseline quality, outperforming both DiCache and
FasterCache in speed--quality trade-off. Our contributions are:
\begin{enumerate}
    \item We formalize feature caching for DiT-based world models
    as a \emph{dynamical approximation} problem and identify the
    zero-order hold assumption in prior methods as the primary
    source of ghosting, blur, and motion incoherence in dynamic
    rollouts.

    \item We introduce WorldCache, a unified framework that
    improves both \emph{when} to skip (motion and saliency-aware
    decisions) and \emph{how} to approximate (optimal blending and
    motion compensation), while adapting to the denoising phase.

    \item We demonstrate state-of-the-art training-free
    acceleration on multiple DiT backbones, achieving up to
    $\mathbf{2.3\times}$ speedup with $\mathbf{99.4\%}$ quality
    retention on Cosmos-Predict2.5, and show that the approach
    transfers across model scales and conditioning modalities.
\end{enumerate}

\section{Related Work}

\paragraph{Video diffusion and world simulators.}
Diffusion models have become a leading approach for high-fidelity video generation, from early formulations \cite{ho2022videodiffusion} to scalable latent/cascaded pipelines \cite{he2022lvdm,ho2022imagenvideo,blattmann2023svd} and large-scale text-to-video systems \cite{singer2022makeavideo}.  
Recently, video generation models have also been studied as \emph{world simulators}, evaluated for physical consistency and action-relevant prediction \cite{openai2024worldsimulators,qin2024worldsimbench}. In this direction, NVIDIA’s Cosmos platform/Cosmos-Predict target physical AI simulation \cite{nvidia2025cosmosplatform,ali2025world}, with benchmarks such as PAI-Bench to assess physical plausibility and controllability \cite{zhou2025paibench}. Related efforts include interactive environment world models \cite{bruce2024genie} and large token-based models for video generation \cite{kondratyuk2024videopoet}.

\paragraph{Efficient diffusion inference.}
A common acceleration axis is reducing sampling cost via fewer or cheaper denoising steps. Training-free methods include alternative samplers such as DDIM \cite{song2020ddim} and fast solvers such as DPM-Solver/DPM-Solver++ \cite{lu2022dpmsolver,lu2022dpmsolverpp}, while distillation compresses many-step teachers into few-step students \cite{salimans2022progressivedistillation}. WorldCache instead keeps the base model and schedule, and reduces compute via safe reuse of internal activations.

\paragraph{Caching and reuse in diffusion transformers.}
Caching methods exploit redundancy across timesteps and guidance passes. DeepCache reuses high-level features across adjacent steps (mainly for U-Nets) \cite{ma2024deepcache}. For video diffusion transformers, \textbf{FasterCache} accelerates inference by reusing attention features across timesteps and introducing a CFG cache that reuses conditional/unconditional redundancy to reduce guidance overhead \cite{lyu2025fastercache}. DeepCache~\cite{ma2024deepcache} shows that reusing high-level features across steps can accelerate diffusion inference. \textbf{DiCache} further makes caching adaptive with an online probe to decide \emph{when} to refresh and a trajectory-aligned reuse strategy to decide \emph{how} to combine cached states \cite{bu2026dicache}. Despite strong gains, caching can still be brittle when motion, fine textures, or semantically important regions cause cached states to drift.

\paragraph{Motion-compensated and perception-aware reuse.}
Feature reuse has also been explored in video recognition via propagation with optical flow \cite{zhu2017deepfeatureflow} and multi-rate update schedules \cite{shelhamer2016clockwork}, motivating alignment-aware reuse rather than fixed-coordinate copying. Classic and modern flow methods (Lucas--Kanade, RAFT) \cite{lucas1981kanade,teed2020raft} illustrate the accuracy/efficiency trade-off for motion compensation. Perceptual quality can be tracked with deep perceptual metrics and structure/texture-aware measures \cite{zhang2018lpips,ding2020dists}, while Laplacian pyramids provide a classical multi-scale view of high-frequency detail \cite{burt1983laplacian}. WorldCache builds on these ideas with motion-aligned reuse, saliency-aware monitoring, and principled temporal extrapolation inspired by system identification \cite{ljung1999systemidentification}.

\section{Method}
\label{sec:method}

\subsection{Preliminaries: DiT Denoising in World Models}
\label{sec:prelim}

We consider a DiT-based world model that predicts future visual
states by iteratively denoising a latent video representation. Let
$\mathbf{z}_t \in \mathbb{R}^{B \times T \times H \times W \times
D}$ denote the latent tensor at denoising step~$t$ (not to be
confused with video frame index), where $B$ is batch size, $T$ is
the number of latent frames, $H \times W$ is spatial resolution,
and $D$ is the channel dimension. The denoiser is a stack of $N$
transformer blocks $\{F_i\}_{i=1}^{N}$, producing
$\mathbf{z}_t^{(i)} = F_i(\mathbf{z}_t^{(i-1)})$, with
$\mathbf{z}_t^{(0)}=\mathbf{z}_t$ and $\mathbf{z}_t^{(N)}$ used
by the sampler to obtain $\mathbf{z}_{t+1}$. Throughout this
section, superscripts in parentheses denote layer indices and
subscripts denote denoising steps.

\subsection{Foundation: Probe-Then-Cache}
\label{sec:dicache_foundation}

WorldCache inherits its architectural skeleton from the
probe-then-cache paradigm introduced by
DiCache~\cite{bu2026dicache}, but \emph{replaces} both its skip
criterion and its reuse mechanism. We first describe the shared
skeleton, then identify the two components we redesign.

\vspace{0.25em}
\noindent\textbf{Probe (inherited).} At each step~$t$, only the
first $k$~blocks (probe depth) are evaluated to obtain
$\mathbf{z}_t^{(k)}$. A drift indicator approximates the
deep-layer change:
\begin{equation}
  \delta_t =
  \frac{\left\| \mathbf{z}_t^{(k)}
        - \mathbf{z}_{t-1}^{(k)} \right\|_1}
       {\left\| \mathbf{z}_{t-1}^{(k)} \right\|_1 + \epsilon}.
  \label{eq:probe_drift}
\end{equation}
If $\delta_t$ falls below a threshold, blocks $k{+}1,\dots,N$ are
skipped and cached deep states are reused; otherwise the full
network executes and the cache is refreshed.

\vspace{0.25em}
\noindent\textbf{Skip criterion (replaced).} DiCache uses a fixed
global threshold on $\delta_t$. WorldCache replaces this with
motion-adaptive, saliency-weighted decisions
(Secs.~\ref{sec:cfc}--\ref{sec:swd}).

\noindent\textbf{Reuse mechanism (replaced).} DiCache estimates a
scalar blending coefficient from L1 residual ratios and
interpolates between cached states from steps $t{-}1$ and
$t{-}2$. This captures the \emph{magnitude} of feature evolution
but discards directional information. WorldCache replaces this
with a vector-projection-based approximation and optional
motion-compensated warping (Sec.~\ref{sec:ofa}).

\begin{figure}[t!]
    \centering
    \includegraphics[width=\textwidth]{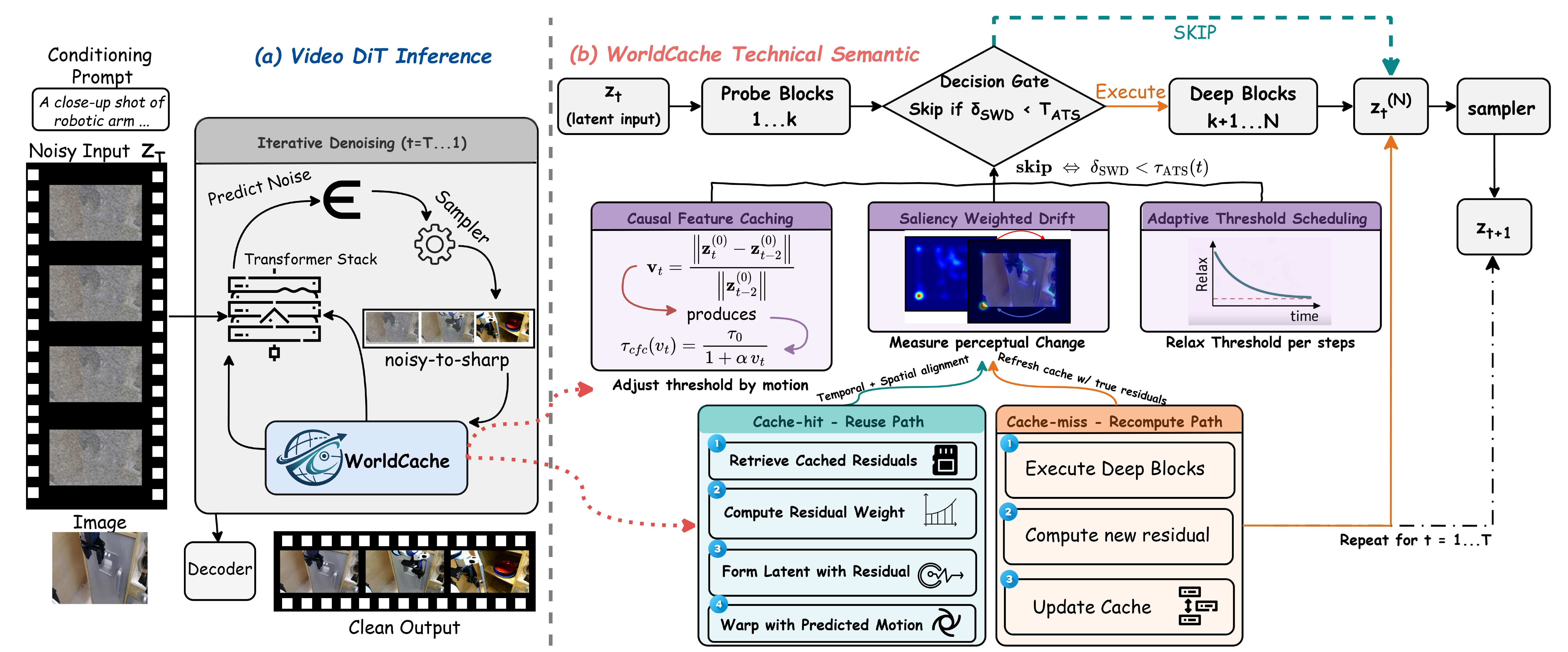}
    \caption{A DiT world model denoises latent video states ($z_t$) by running probe blocks (1$\ldots$k) followed by deep blocks ($k{+}1\ldots(N)$). WorldCache inserts a decision gate that skips deep blocks when the saliency-weighted probe drift ($\delta_{\mathrm{SWD}}$) is below a motion and step-adaptive threshold ($\tau_{\mathrm{ATS}}(t)$) (computed by CFC+ATS), enabling cache hits. On a cache hit, OFA reuses computation by aligning cached residuals via optimal interpolation and optional motion-compensated warping to approximate ($z_t^{(N)}$). On a cache miss, the model executes deep blocks, updates the residual cache in a ping-pong buffer, and continues the denoising loop for (t=1$\ldots$T).}
    \label{fig:method}
\end{figure}

\subsection{WorldCache Overview}
\label{sec:worldcache_overview}

Fig.~\ref{fig:method} summarizes the full pipeline. At each
denoising step~$t$, the probe computes shallow features
$\mathbf{z}_t^{(k)}$. \textbf{CFC} (Sec.~\ref{sec:cfc}) and
\textbf{SWD} (Sec.~\ref{sec:swd}) jointly determine whether to
skip by combining a motion-adaptive threshold with a
saliency-weighted drift signal. On a cache hit, \textbf{OFA}
(Sec.~\ref{sec:ofa}) approximates the deep output via
least-squares optimal blending and optional spatial warping.
\textbf{ATS} (Sec.~\ref{sec:ats}) modulates the skip threshold
across the denoising trajectory, tightening it during
structure-formation steps and relaxing it during late refinement.
All four modules are training-free and add negligible overhead to
the probe computation.

\subsection{Causal Feature Caching (CFC): Motion-Adaptive
Decisions}
\label{sec:cfc}

\textbf{\textit{When is reuse safe?}}
In world-model video, the amount of motion varies substantially
across prompts and across denoising steps. A fixed threshold is
overly permissive during fast motion (risking ghosting) and overly
conservative during static intervals (missing speedups). CFC
adapts the skip threshold using an inexpensive motion proxy
derived from the raw latent input. We define a ``velocity'' as the
normalized two-step input change:
\begin{equation}
  v_t =
  \frac{\left\| \mathbf{z}_t^{(0)}
        - \mathbf{z}_{t-2}^{(0)} \right\|_1}
       {\left\| \mathbf{z}_{t-2}^{(0)} \right\|_1 + \epsilon}.
  \label{eq:cfc_velocity}
\end{equation}
We use a two-step gap because step $t{-}1$ may itself be a cached
approximation; anchoring to $t{-}2$ (the most recent
fully-computed input) yields a more reliable velocity estimate. The
motion-adaptive threshold is:
\begin{equation}
  \tau_{\text{CFC}}(v_t)
  = \frac{\tau_0}{1 + \alpha \cdot v_t},
  \label{eq:cfc_tau}
\end{equation}
where $\tau_0$ is the base threshold and $\alpha$ controls
sensitivity. When dynamics are fast ($v_t$ large),
$\tau_{\text{CFC}}$ tightens, making skips less likely; when
dynamics are slow, $\tau_{\text{CFC}} \approx \tau_0$. We maintain
a ping-pong buffer (two alternating cache slots indexed by step
parity) so that reuse is always anchored to one of the two most
recent fully-computed states.

\begin{figure}[t]
    \centering
    \includegraphics[width=\linewidth,trim=0 0 0 30,clip]{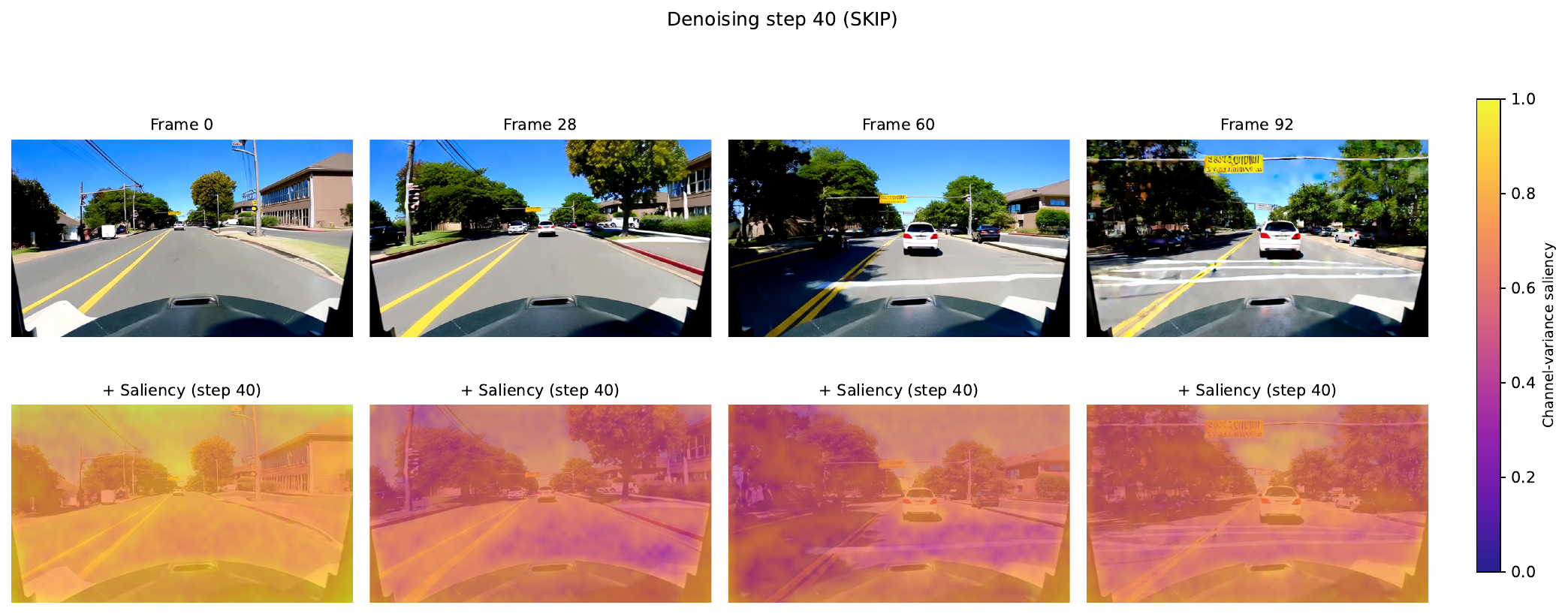}
    \caption{\textbf{Saliency overlay (Cosmos-Predict2.5, 2B).}
    Channel-variance saliency from step~40 overlaid on four video frames. High saliency (yellow) marks structurally complex regions where caching errors are most visible; low saliency (purple) marks smooth areas where reuse is safe. SWD reweights drift using this signal to prioritize fidelity on detail-rich content.}
    \label{fig:saliency_overlay}
\end{figure}

\subsection{Saliency-Weighted Drift (SWD): Perception-Aware
Probing}
\label{sec:swd}

\textbf{\textit{Is the drift signal measuring the right thing?}}
The global drift $\delta_t$ (Eq.~\ref{eq:probe_drift}) treats
every spatial location equally, so it cannot distinguish between
harmless background fluctuation and critical foreground change.
SWD reweights drift toward perceptually important regions,
ensuring that the method recomputes when salient content changes
and skips when only the background drifts.

We define a spatial saliency map from the channel-wise variance of
probe features:
\begin{equation}
  S_{h,w}
  = \operatorname{Var}_d\!\!\left(
      \bar{\mathbf{z}}_t^{(k)}[h, w, :]
    \right),
  \label{eq:swd_saliency}
\end{equation}
where $\bar{\mathbf{z}}_t^{(k)}$ is the probe output averaged over
the batch and temporal axes, and the variance is taken over the
channel dimension~$d$. High channel variance indicates spatially
complex, information-rich regions (edges, textures, object
boundaries) where caching errors are most perceptually
visible (Fig.~\ref{fig:saliency_overlay}). We normalize
to $\hat{S} \in [0,1]$ and define the saliency-weighted drift:
\begin{equation}
  \delta_t^{\text{SWD}}
  = \frac{1}{HW} \sum_{h,w}
    \left\| \mathbf{z}_t^{(k)}(h,w)
           - \mathbf{z}_{t-1}^{(k)}(h,w)
    \right\|_1
    \cdot \left(1 + \beta_s\, \hat{S}_{h,w}\right),
  \label{eq:swd_drift}
\end{equation}
where $\beta_s$ controls saliency emphasis. The weighting term
$(1 + \beta_s\, \hat{S}_{h,w})$ amplifies drift contributions from
salient regions and attenuates those from featureless backgrounds.
Consequently, a scene where only the static sky changes produces a
low $\delta_t^{\text{SWD}}$ (safe to skip), while one where a
foreground agent moves, even slightly, produces a high
$\delta_t^{\text{SWD}}$ (triggering recomputation). The final skip
decision combines SWD with the motion-adaptive threshold from CFC:
\begin{equation}
  \text{skip}
  \iff
  \delta_t^{\text{SWD}} < \tau_{\text{CFC}}(v_t).
  \label{eq:swd_decision}
\end{equation}

\subsection{Optimal Feature Approximation (OFA): Improved
Reuse Quality}
\label{sec:ofa}

\textbf{\textit{When we skip, can we produce a better
approximation?}}
CFC and SWD decide \emph{when} to skip. OFA improves \emph{what}
is produced on a cache hit, via two complementary operators: one
temporal (least-squares optimal blending) and one spatial
(motion-compensated warping).

\subsubsection{Optimal State Interpolation (OSI)}
\label{sec:osi}

On a cache hit, the deep output $\mathbf{z}_t^{(N)}$ must be
approximated from cached history. DiCache~\cite{bu2026dicache}
estimates a scalar blending coefficient $\gamma$ from L1 distance
ratios between probe residuals. This captures the magnitude of
feature evolution but discards directional information: when motion
causes the feature trajectory to curve, the scalar ratio
extrapolates along a stale direction, and the resulting errors
accumulate over consecutive cache hits.

We reformulate the estimation as a least-squares vector
projection. Define the deep computation residual:
\begin{equation}
  \mathbf{r}_t = \mathbf{z}_t^{(N)} - \mathbf{z}_t^{(0)},
  \label{eq:residual}
\end{equation}
and on a cache hit, let $\tilde{\mathbf{r}}_t =
\mathbf{z}_t^{(k)} - \mathbf{z}_t^{(0)}$ be the probe-derived
partial residual. We seek a gain $\gamma^*$ that best aligns the
recent residual trajectory with the current probe signal:
\begin{equation}
  \Delta_{\text{tgt}} = \tilde{\mathbf{r}}_t - \mathbf{r}_{t-2},
  \qquad
  \Delta_{\text{src}} = \mathbf{r}_{t-1} - \mathbf{r}_{t-2},
\end{equation}
\begin{equation}
  \gamma^*
  = \arg\min_{\gamma}
    \left\| \Delta_{\text{tgt}}
           - \gamma\,\Delta_{\text{src}} \right\|^2
  = \frac{\langle \Delta_{\text{tgt}},\,
                   \Delta_{\text{src}} \rangle}
         {\|\Delta_{\text{src}}\|^2 + \epsilon}.
  \label{eq:osi_gamma}
\end{equation}
We clamp $\gamma^*$ to $[0, \gamma_{\max}]$ (we use
$\gamma_{\max}{=}2$) to prevent blow-up when
$\|\Delta_{\text{src}}\|$ is small. The deep output is
approximated as:
\begin{equation}
  \hat{\mathbf{z}}_t^{(N)}
  = \mathbf{z}_t^{(0)}
    + \mathbf{r}_{t-2}
    + \gamma^*\!\left(\mathbf{r}_{t-1} - \mathbf{r}_{t-2}\right).
  \label{eq:osi_out}
\end{equation}
The inner product in Eq.~\ref{eq:osi_gamma} is the key difference
from scalar-ratio methods. When the feature trajectory curves
(\eg, a moving object changes direction), the dot product
naturally attenuates $\gamma^*$, preventing extrapolation along a
stale direction. When the trajectory is linear, OSI recovers the
same estimate as scalar-ratio methods. OSI thus generalizes
scalar-ratio alignment; we verify the improvement empirically in
the ablation study.

\subsubsection{Motion-Compensated Feature Warping}
\label{sec:warp}

OSI corrects \emph{temporal} misalignment in the residual
trajectory, but cached features from step $t{-}1$ may also be
\emph{spatially} misaligned when the scene contains motion. OFA
optionally warps cached features to the current coordinate frame
before applying OSI.

We estimate a displacement field between consecutive latent inputs
via multi-scale correlation in latent space (no external network):
\begin{equation}
  \mathbf{u}_{t \to t-1}
  = \operatorname{LatentCorr}\!\left(
      \mathbf{z}_t^{(0)},\, \mathbf{z}_{t-1}^{(0)}
    \right),
  \label{eq:ofa_flow}
\end{equation}
which adds less than 3\% overhead per cached step. The cached deep
features are then warped:
\begin{equation}
  \tilde{\mathbf{z}}_{t-1}^{(N)}
  = \operatorname{Warp}\!\left(
      \mathbf{z}_{t-1}^{(N)},\, \mathbf{u}_{t \to t-1}
    \right),
  \label{eq:ofa_warp}
\end{equation}
and $\tilde{\mathbf{z}}_{t-1}^{(N)}$ replaces
$\mathbf{z}_{t-1}^{(N)}$ in the residual computation of
Eq.~\ref{eq:osi_out}. That is, OSI operates on the
spatially-corrected residuals $\tilde{\mathbf{r}}_{t-1} =
\tilde{\mathbf{z}}_{t-1}^{(N)} - \mathbf{z}_{t-1}^{(0)}$,
reducing compound spatial drift that is especially harmful in
autoregressive world-model rollouts. We disable warping during the
first five denoising steps, where the low signal-to-noise ratio
makes displacement estimation unreliable.

\begin{figure}[t!]
    \centering
    \includegraphics[width=\textwidth]{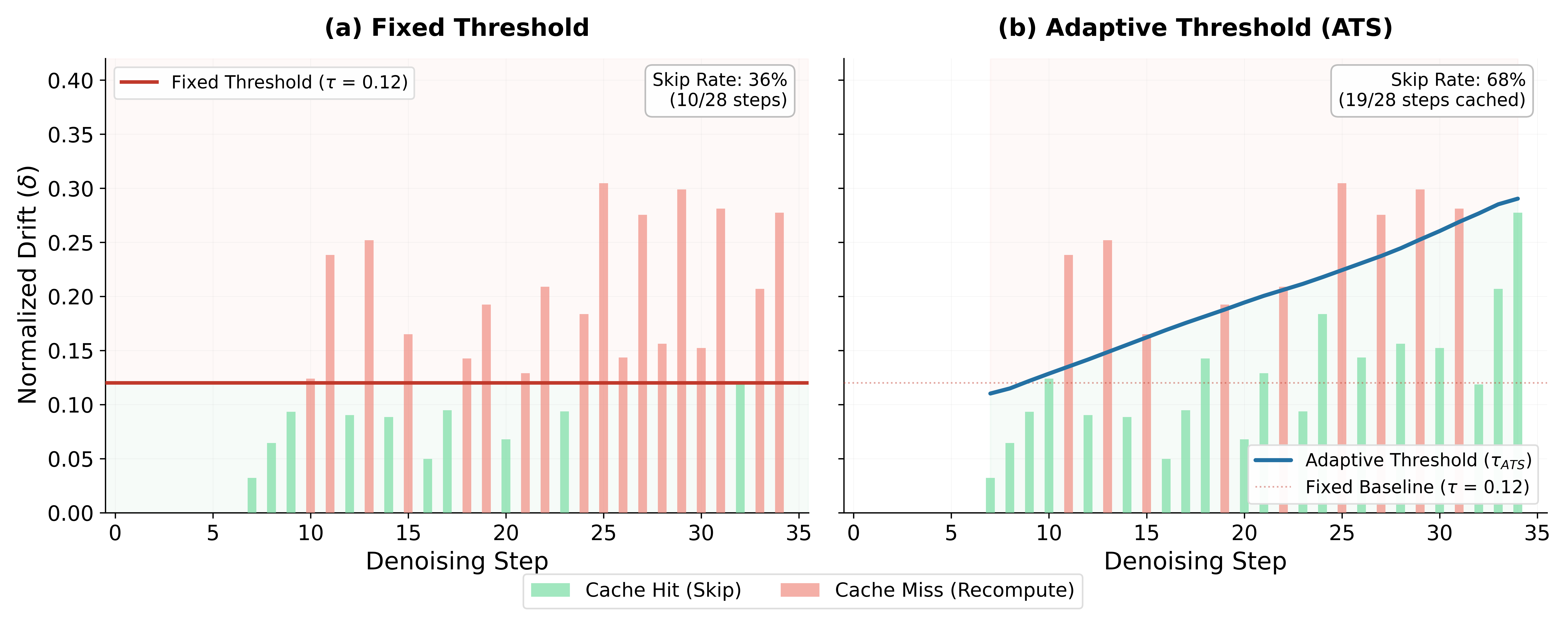}
    \caption{\textbf{Comparison of static vs. adaptive caching strategies}. (a) A Fixed Threshold ($\tau=0.12$) fails to accommodate the naturally increasing drift ($\delta$) in later denoising steps, resulting in frequent recomputations and a low skip rate (36\%). (b) Our Adaptive Threshold (ATS) dynamically scales $\tau_{ATS}$ with the expected drift. This allows ATS to capture significantly more cache hits (green bars) in later stages, nearly doubling the overall skip rate (68\%) while maintaining generation quality.}
    \label{fig:ATS}
\end{figure}

\begin{table}[t]
\centering
\caption{\textbf{Text2World (T2W) generation results on PAI-Bench across two model scales.} We evaluate four methods, Baseline (no acceleration), DiCache, FasterCache, and WorldCache (ours), on Cosmos-Predict2.5 at both 2B and 14B parameter scales. Domain Score aggregates seven semantic categories (CS: City Street, AV: Aerial View, RO: Road, IN: Indoor, HU: Human, PH: Physics, MI: Mixed), while Quality Score spans eight perceptual and fidelity dimensions (SC: Scene Consistency, BC: Background Consistency, MS: Motion Smoothness, AQ: Aesthetic Quality, IQ: Image Quality, OC: Object Consistency, IS: Imaging Subject, IB: Imaging Background). Avg.\ denotes the overall score averaged across both Domain and Quality metrics. Lat.\ reports wall-clock inference latency (seconds), and Speedup is relative to the unaccelerated Baseline. $\Delta$ rows quantify the per-metric gain of WorldCache over DiCache, the strongest competing method. WorldCache achieves the best accuracy–efficiency trade-off, consistently outperforming DiCache while delivering up to $\mathbf{2.10\times}$ speedup at 2B and $\mathbf{2.14\times}$ at 14B with negligible quality degradation.}
\label{tab:main_results_t2w_fine}
\vspace{2pt}
\setlength{\tabcolsep}{2.1pt}
\renewcommand{\arraystretch}{1.08}
\scriptsize
\resizebox{\textwidth}{!}{%
\begin{tabular}{l | ccccccc | c | cccccccc | c | c cc}
\toprule
& \multicolumn{8}{c|}{\textbf{Domain Score}} 
& \multicolumn{9}{c|}{\textbf{Quality Score}}
& \makecell{\textbf{Avg.}\\$\uparrow$}
& \makecell{\textbf{Lat. (s)}\\$\downarrow$}
& \makecell{\textbf{Speedup}\\$\uparrow$} \\
\cmidrule(lr){2-9}\cmidrule(lr){10-18}
\textbf{Method} 
& CS & AV & RO & IN & HU & PH & MI & \makecell{\textbf{Avg}\\$\uparrow$}
& SC & BC & MS & AQ & IQ & OC & IS & IB & \makecell{\textbf{Avg}\\$\uparrow$}
& & & \\
\midrule
\rowcolor{orange!10}
\multicolumn{21}{c}{\textit{Cosmos-Predict2.5 -- 2B}} \\
\midrule
Baseline
& 0.759 & 0.643 & 0.724 & 0.820 & 0.769 & 0.859 & 0.846 & 0.767
& 0.909 & 0.929 & 0.979 & 0.501 & 0.712 & 0.199 & 0.788 & 0.808 & 0.728
& 0.748 & 54.34 & 1.0$\times$ \\
DiCache
& 0.756 & 0.631 & 0.707 & 0.799 & 0.773 & 0.849 & 0.833 & 0.759
& 0.902 & 0.925 & 0.978 & 0.493 & 0.705 & 0.197 & 0.780 & 0.838 & 0.727
& 0.743 & 40.82 & 1.3$\times$ \\
FasterCache
& 0.675 & 0.553 & 0.549 & 0.691 & 0.652 & 0.719 & 0.745 & 0.629
& 0.849 & 0.909 & 0.970 & 0.405 & 0.594 & 0.176 & 0.709 & 0.796 & 0.676
& 0.652 & 34.51 & 1.6$\times$ \\
WorldCache
& 0.759 & 0.639 & 0.735 & 0.810 & 0.760 & 0.845 & 0.839 & \textbf{0.763}
& 0.903 & 0.927 & 0.979 & 0.492 & 0.703 & 0.196 & 0.782 & 0.826 & \textbf{0.727}
& \textbf{0.745} & \textbf{26.28} & \textbf{2.1$\times$} \\
\midrule
\rowcolor{gray!10}
\textit{$\Delta$ (WC$-$DC)}
& +0.003 & +0.008 & +0.028 & +0.011 & $-$0.013 & $-$0.004 & +0.006 & +0.004
& +0.001 & +0.002 & +0.001 & $-$0.001 & $-$0.002 & $-$0.001 & +0.002 & $-$0.012 & +0.000
& +0.002 & $-$14.54 & +0.80$\times$ \\
\midrule
\rowcolor{green!10}
\multicolumn{21}{c}{\textit{Cosmos-Predict2.5 -- 14B}} \\
\midrule
Baseline
& 0.782 & 0.643 & 0.762 & 0.828 & 0.794 & 0.900 & 0.880 & 0.792
& 0.940 & 0.948 & 0.988 & 0.518 & 0.719 & 0.202 & 0.806 & 0.846 & 0.746
& 0.769 & 216.25 & 1.0$\times$ \\
DiCache
& 0.795 & 0.645 & 0.757 & 0.819 & 0.790 & 0.906 & 0.880 & 0.792
& 0.939 & 0.949 & 0.988 & 0.518 & 0.714 & 0.201 & 0.806 & 0.845 & 0.745
& 0.768 & 148.36 & 1.4$\times$ \\
FasterCache
& 0.707 & 0.564 & 0.584 & 0.710 & 0.677 & 0.773 & 0.785 & 0.659
& 0.884 & 0.930 & 0.979 & 0.427 & 0.604 & 0.180 & 0.731 & 0.821 & 0.694
& 0.676 & 126.60 & 1.7$\times$ \\
WorldCache
& 0.792 & 0.659 & 0.751 & 0.838 & 0.794 & 0.908 & 0.879 & \textbf{0.795}
& 0.940 & 0.948 & 0.987 & 0.517 & 0.718 & 0.201 & 0.804 & 0.856 & \textbf{0.746}
& \textbf{0.771} & \textbf{98.61} & \textbf{2.14$\times$} \\
\midrule
\rowcolor{gray!10}
\textit{$\Delta$ (WC$-$DC)}
& $-$0.003 & +0.014 & $-$0.006 & +0.019 & +0.004 & +0.002 & $-$0.001 & +0.003
& +0.001 & $-$0.001 & $-$0.001 & $-$0.001 & +0.004 & +0.000 & $-$0.002 & +0.011 & +0.001
& +0.003 & $-$49.75 & +0.74$\times$ \\
\bottomrule
\end{tabular}%
}
\end{table}

\subsection{Adaptive Threshold Scheduling (ATS): Phase-Aware
Reuse}
\label{sec:ats}

\textbf{\textit{Can we push acceleration further without breaking
fidelity?}}
The preceding modules (CFC, SWD, OFA) establish a perception-aware
caching infrastructure, but they operate with a fixed base
threshold $\tau_0$. The denoising trajectory, however, has two
distinct phases. During \emph{structure formation} (early steps,
high noise), the network makes large, semantically critical
updates that establish global layout and motion. During
\emph{detail refinement} (late steps, low noise), updates become
small, high-frequency corrections. A static threshold calibrated
for the early phase becomes unnecessarily conservative in the late
phase: empirically, the cache hit rate drops sharply after step~20
(out of~35) because probe drift falls below the detection floor
while the threshold remains unchanged (Fig.~\ref{fig:ATS}). ATS addresses this mismatch with a step-dependent relaxation:
\begin{equation}
  \tau_{\text{ATS}}(t)
  = \tau_{\text{CFC}}(v_t)
    \cdot \left(1 + \beta_d \cdot \frac{t}{T}\right),
  \label{eq:ats_tau}
\end{equation}
where $t \in [0, T]$ is the denoising step, $T$ is the total
number of steps, and $\beta_d$ controls the relaxation rate. At an
early step (\eg, $t{=}2$, $T{=}35$, $\beta_d{=}4.0$), the
multiplier is ${\approx}1.2$, keeping the threshold tight and
forcing full execution for structure-critical updates. At a late
step (\eg, $t{=}32$), the multiplier reaches ${\approx}4.6$,
aggressively relaxing the threshold. Since the scene geometry is
already established and the network produces only fine texture
corrections, the cached OFA approximation is safely reused across
consecutive steps. This aggressive late-stage relaxation is the
primary source of WorldCache's speedup gains. The ablation in
Sec.~\ref{sec:ablations} confirms that it reduces quality by less
than 0.6\% relative to the baseline.

\section{Experiments}
\label{sec:experiments}

\subsection{Experimental Setup}

\noindent \textbf{\textit{Base models.}}
We evaluate WorldCache on the Cosmos-Predict2.5 family of Video World Models~\cite{ali2025world} at two scales: the 2B-parameter and 14B-parameter variants. Both models employ a Video Diffusion Transformer backbone with 3D Rotary Positional Encoding (RoPE) and are trained with a Flow Matching (velocity prediction) objective. Each sampling run generates 93 frames (${\sim}5.8$s at 16\,FPS), corresponding to 24 latent frames, using 35 denoising steps with Euler scheduling. To demonstrate that WorldCache is not specific to a single backbone, we additionally evaluate on \textbf{WAN2.1}\cite{wan2025wan}, a DiT-based video generation model, using the official inference configuration for the corresponding checkpoints. 

\noindent \textbf{\textit{Benchmark.}}
All methods are evaluated on PAI-Bench (Physical AI Benchmark)~\cite{zhou2025paibench}, a comprehensive evaluation suite designed specifically for Video World Models. PAI-Bench spans six physical domains, Robot, Autonomous Vehicles, Human Activity, Industry, Common Sense, and Physics, and reports a \textit{Domain Score} (physical plausibility), a \textit{Quality Score} (visual fidelity), and their average as the \textit{Overall Score}. We report results on both the \textbf{Text-to-World (T2W)} and \textbf{Image-to-World (I2W)} generation tasks. More details on each evaluation metric are provided in supplementary material. 

\noindent \textbf{\textit{Baselines.}}
We compare against two state-of-the-art training-free caching baselines: \textbf{DiCache}~\cite{bu2026dicache}, which employs an online probe profiling scheme with trajectory-aligned residual reuse; and \textbf{FasterCache}~\cite{lyu2025fastercache}, which uses a fixed step-skipping schedule combined with CFG-Cache for unconditional branch reuse. All methods are applied as drop-in replacements on the same base model checkpoints. Wall-clock latency is measured on a single NVIDIA H200 GPU with identical batch size and precision settings.

\noindent \textbf{\textit{WorldCache configuration.}}
Unless otherwise stated, the full WorldCache pipeline combines four modules: Causal Feature Caching (CFC), Saliency-Weighted Drift (SWD), Optimal Feature Approximation (OFA), and Adaptive Threshold Scheduling (ATS). We set the base threshold $\tau_0=0.08$, motion sensitivity $\alpha=2.0$, saliency weight $\beta_s=0.12$, and $\beta_d{=}4.0$ across the diffusion trajectory. All hyperparameters are fixed across models and tasks without per-prompt tuning. Unless otherwise stated, we also apply the same caching pipeline and hyperparameter selection protocol to WAN2.1 as in Cosmos, and report results on the same PAI-Bench\cite{zhou2025paibench} tasks for direct comparison.

\subsection{Main Results}

\begin{table}[t]
\centering
\caption{\textbf{Image2World (I2W) generation results on PAI-Bench across two model scales.} We evaluate four methods: Baseline (no acceleration), DiCache, FasterCache, and WorldCache (ours) on Cosmos-Predict2.5 at both 2B and 14B parameter scales under the Image2World setting, where a conditioning image guides video world generation. $\Delta$ rows quantify the per-metric gain of WorldCache over DiCache. Compared to the T2W setting, I2W scores are generally higher owing to the additional visual grounding provided by the conditioning image. WorldCache achieves the best accuracy--efficiency trade-off, consistently outperforming DiCache while delivering up to $\mathbf{2.3\times}$ speedup at 2B and $\mathbf{2.18\times}$ at 14B with negligible quality degradation across all domain and quality dimensions.}

\label{tab:main_results_i2w_fine}
\setlength{\tabcolsep}{2.0pt}
\renewcommand{\arraystretch}{1.08}
\scriptsize
\resizebox{\textwidth}{!}{%
\begin{tabular}{l | ccccccc | c | cccccccc | c | c cc}
\toprule
& \multicolumn{8}{c|}{\textbf{Domain Score}} 
& \multicolumn{9}{c|}{\textbf{Quality Score}}
& \makecell{\textbf{Avg.}\\$\uparrow$}
& \makecell{\textbf{Lat. (s)}\\$\downarrow$}
& \makecell{\textbf{Speedup}\\$\uparrow$} \\
\cmidrule(lr){2-9}\cmidrule(lr){10-18}
\textbf{Method} 
& CS & AV & RO & IN & HU & PH & MI & \makecell{\textbf{Avg}\\$\uparrow$}
& SC & BC & MS & AQ & IQ & OC & IS & IB & \makecell{\textbf{Avg}\\$\uparrow$}
& & & \\
\midrule
\rowcolor{orange!10}
\multicolumn{21}{c}{\textit{Cosmos-Predict2.5 -- 2B}} \\
\midrule
Baseline
& 0.919 & 0.694 & 0.811 & 0.877 & 0.840 & 0.909 & 0.886 & 0.845
& 0.896 & 0.929 & 0.982 & 0.505 & 0.674 & 0.212 & 0.936 & 0.952 & 0.761
& 0.803 & 55.04 & 1.0$\times$ \\
DiCache
& 0.899 & 0.697 & 0.791 & 0.876 & 0.828 & 0.887 & 0.909 & 0.835
& 0.885 & 0.923 & 0.980 & 0.492 & 0.660 & 0.212 & 0.927 & 0.940 & 0.752
& 0.794 & 39.68 & 1.4$\times$ \\
FasterCache
& 0.855 & 0.676 & 0.697 & 0.829 & 0.739 & 0.851 & 0.847 & 0.772
& 0.800 & 0.872 & 0.974 & 0.432 & 0.577 & 0.197 & 0.888 & 0.919 & 0.708
& 0.740 & 32.75 & 1.7$\times$ \\
WorldCache
& 0.912 & 0.708 & 0.796 & 0.876 & 0.833 & 0.893 & 0.890 & \textbf{0.840}
& 0.892 & 0.926 & 0.982 & 0.496 & 0.661 & 0.212 & 0.931 & 0.948 & \textbf{0.756}
& \textbf{0.798} & \textbf{24.48} & \textbf{2.3$\times$} \\
\midrule
\rowcolor{gray!10}
\textit{$\Delta$ (WC$-$DC)}
& +0.013 & +0.011 & +0.005 & +0.000 & +0.005 & +0.006 & $-$0.019 & +0.005
& +0.007 & +0.003 & +0.002 & +0.004 & +0.001 & +0.000 & +0.004 & +0.008 & +0.004
& +0.004 & $-$15.20 & +0.9$\times$ \\
\midrule
\rowcolor{green!10}
\multicolumn{21}{c}{\textit{Cosmos-Predict2.5 -- 14B}} \\
\midrule
Baseline
& 0.920 & 0.716 & 0.826 & 0.905 & 0.849 & 0.922 & 0.924 & 0.860
& 0.912 & 0.935 & 0.988 & 0.510 & 0.665 & 0.213 & 0.958 & 0.966 & 0.769
& 0.814 & 210.07 & 1.0$\times$ \\
DiCache
& 0.913 & 0.716 & 0.826 & 0.886 & 0.844 & 0.920 & 0.921 & 0.855
& 0.911 & 0.935 & 0.988 & 0.509 & 0.658 & 0.212 & 0.956 & 0.965 & 0.767
& 0.811 & 146.04 & 1.4$\times$ \\
FasterCache
& 0.856 & 0.688 & 0.715 & 0.842 & 0.743 & 0.869 & 0.862 & 0.782
& 0.813 & 0.873 & 0.975 & 0.437 & 0.567 & 0.195 & 0.906 & 0.930 & 0.712
& 0.747 & 123.75 & 1.7$\times$ \\
WorldCache
& 0.923 & 0.727 & 0.824 & 0.901 & 0.845 & 0.925 & 0.909 & \textbf{0.859}
& 0.912 & 0.935 & 0.988 & 0.509 & 0.664 & 0.213 & 0.957 & 0.966 & \textbf{0.768}
& \textbf{0.813} & \textbf{99.25} & \textbf{2.18$\times$} \\
\midrule
\rowcolor{gray!10}
\textit{$\Delta$ (WC$-$DC)}
& +0.010 & +0.011 & $-$0.002 & +0.015 & +0.001 & +0.005 & $-$0.012 & +0.004
& +0.001 & +0.000 & +0.000 & +0.000 & +0.006 & +0.001 & +0.001 & +0.001 & +0.001
& +0.002 & $-$46.79 & +0.74$\times$ \\
\bottomrule
\end{tabular}%
}
\end{table}



\noindent \textbf{\textit{Text2World (T2W) on Cosmos-Predict2.5.}}
Table~\ref{tab:main_results_t2w_fine} shows that WorldCache provides the strongest accuracy--efficiency trade-off at both scales.
On \textbf{Cosmos-2B}, WorldCache reduces latency from 54.34\,s to 26.28\,s (\textbf{2.1$\times$}) while preserving the overall average score (0.745 vs.\ 0.748; \mbox{$\sim$99.6\%} retention). DiCache is substantially less aggressive (40.82\,s, 1.3$\times$) and slightly lower in Avg.\ (0.743), while FasterCache is faster than DiCache (34.51\,s, 1.6$\times$) but suffers a large quality drop (Avg.\ 0.652), reflecting degraded world-model fidelity. The $\Delta$ row highlights that WorldCache improves over DiCache most clearly in the Domain categories tied to dynamic scene structure (e.g., RO and IN) while maintaining essentially identical Quality Avg.\ (0.727).
On \textbf{Cosmos-14B}, WorldCache again achieves the best frontier point, reaching 98.61\,s (\textbf{2.14$\times$}) and improving Avg.\ to 0.771 compared to 0.769 for the unaccelerated baseline. In contrast, DiCache achieves 1.4$\times$ (148.36\,s) with Avg.\ 0.768, and FasterCache reaches 1.7$\times$ but with substantially lower Avg.\ (0.676).

\noindent \textbf{\textit{Image2World (I2W) on Cosmos-Predict2.5.}}
As shown in Table~\ref{tab:main_results_i2w_fine}, I2W scores are higher overall due to additional visual grounding, and WorldCache remains the best speed--quality trade-off.
On \textbf{Cosmos-2B}, WorldCache achieves \textbf{2.3$\times$} speedup (55.04\,s $\rightarrow$ 24.48\,s) with Avg.\ 0.798, close to the baseline 0.803. DiCache is slower (1.4$\times$, 39.68\,s) and lower in Avg.\ (0.794), while FasterCache is faster (1.7$\times$) but drops sharply in Avg.\ (0.740). The $\Delta$ row indicates that WorldCache improves over DiCache across most Domain categories and consistently boosts perceptual consistency metrics (SC/BC/MS) while maintaining similar OC.
On \textbf{Cosmos-14B}, WorldCache delivers \textbf{2.18$\times$} speedup (210.07\,s $\rightarrow$ 99.25\,s) with negligible change in Avg.\ relative to baseline (0.813 vs.\ 0.814), outperforming DiCache (1.4$\times$, Avg.\ 0.811) and FasterCache (1.7$\times$, Avg.\ 0.747).

\noindent \textbf{\textit{Transfer to WAN2.1.}}
Table~\ref{tab:wan21_main_single} demonstrates that the benefits of WorldCache transfer beyond Cosmos to WAN2.1.
On \textbf{WAN2.1-1.3B (T2W)}, WorldCache improves over DiCache in both efficiency and score: 61.57\,s $\rightarrow$ 50.84\,s (1.96$\times \rightarrow$ \textbf{2.36$\times$}) while increasing overall score from 0.7703 to 0.7721. 
On \textbf{WAN2.1-14B (I2W)}, WorldCache yields a large latency reduction over DiCache (291.91\,s $\rightarrow$ 206.73\,s; 1.53$\times \rightarrow$ \textbf{2.31$\times$}) and recovers overall score to 0.7388, essentially even surpassing the baseline 0.7384. 
These results confirm that WorldCache consistently improves the DiCache frontier, trading lower fidelity for substantially greater speed across both conditioning modalities and model families. More results on the EgoDex benchmark~\cite{hoque2505egodex} are included in the supplementary material.

\begin{table}[t]
\centering
\caption{\textbf{WAN2.1 results on PAI-Bench.} We report Domain Score (Avg), Quality Score (Avg), Overall Score (computed as the mean of Domain Avg and Quality Avg), wall-clock latency (seconds), and speedup relative to the unaccelerated baseline. Best accelerated results per task/column are in \textbf{bold}.}
\label{tab:wan21_main_single}
\vspace{3pt}
\setlength{\tabcolsep}{3.6pt}
\renewcommand{\arraystretch}{1.10}
\small
\resizebox{0.9\columnwidth}{!}{%
\begin{tabular}{ll ccc cc}
\toprule
\textbf{Task} & \textbf{Method} & \textbf{Domain Avg} $\uparrow$ & \textbf{Quality Avg} $\uparrow$ & \textbf{Overall} $\uparrow$ & \textbf{Latency (s)} $\downarrow$ & \textbf{Speedup} $\uparrow$ \\
\midrule
\rowcolor{orange!10}
\multicolumn{7}{c}{\textit{WAN2.1 - 1.3B}} \\
\midrule
\multirow{4}{*}{T2W}
& Baseline                   & 0.7862 & 0.7592 & 0.7727 & 120.04 & 1.00$\times$ \\
& DiCache                    & 0.7841 & 0.7564 & 0.7703 & 61.57  & 1.96$\times$ \\
& \textbf{WorldCache (Ours)} & \textbf{0.7853} & \textbf{0.7589} & \textbf{0.7721} & \textbf{50.84} & \textbf{2.36$\times$} \\
\midrule
\rowcolor{gray!10}
& \textit{$\Delta$ (WC$-$DC)} & +0.0012 & +0.0025 & +0.0018 & $-$10.73 & +0.40$\times$ \\
\midrule
\rowcolor{green!10}
\multicolumn{7}{c}{\textit{WAN2.1 - 14B}} \\
\midrule
\multirow{4}{*}{I2W}
& Baseline                   & 0.7065 & 0.7703 & 0.7384 & 475.60 & 1.00$\times$ \\
& DiCache                    & 0.6949 & 0.7672 & 0.7311 & 291.91 & 1.53$\times$ \\
& \textbf{WorldCache (Ours)} & \textbf{0.7069} & \textbf{0.7707} & \textbf{0.7388} & \textbf{206.73} & \textbf{2.31$\times$} \\
\midrule
\rowcolor{gray!10}
& \textit{$\Delta$ (WC$-$DC)} & +0.0120 & +0.0035 & +0.0077 & $-$85.18 & +0.78$\times$ \\
\bottomrule
\end{tabular}%
}
\vspace{-0.2in}
\end{table}

\begin{figure}[t!]
    \centering
    \includegraphics[width=\linewidth]{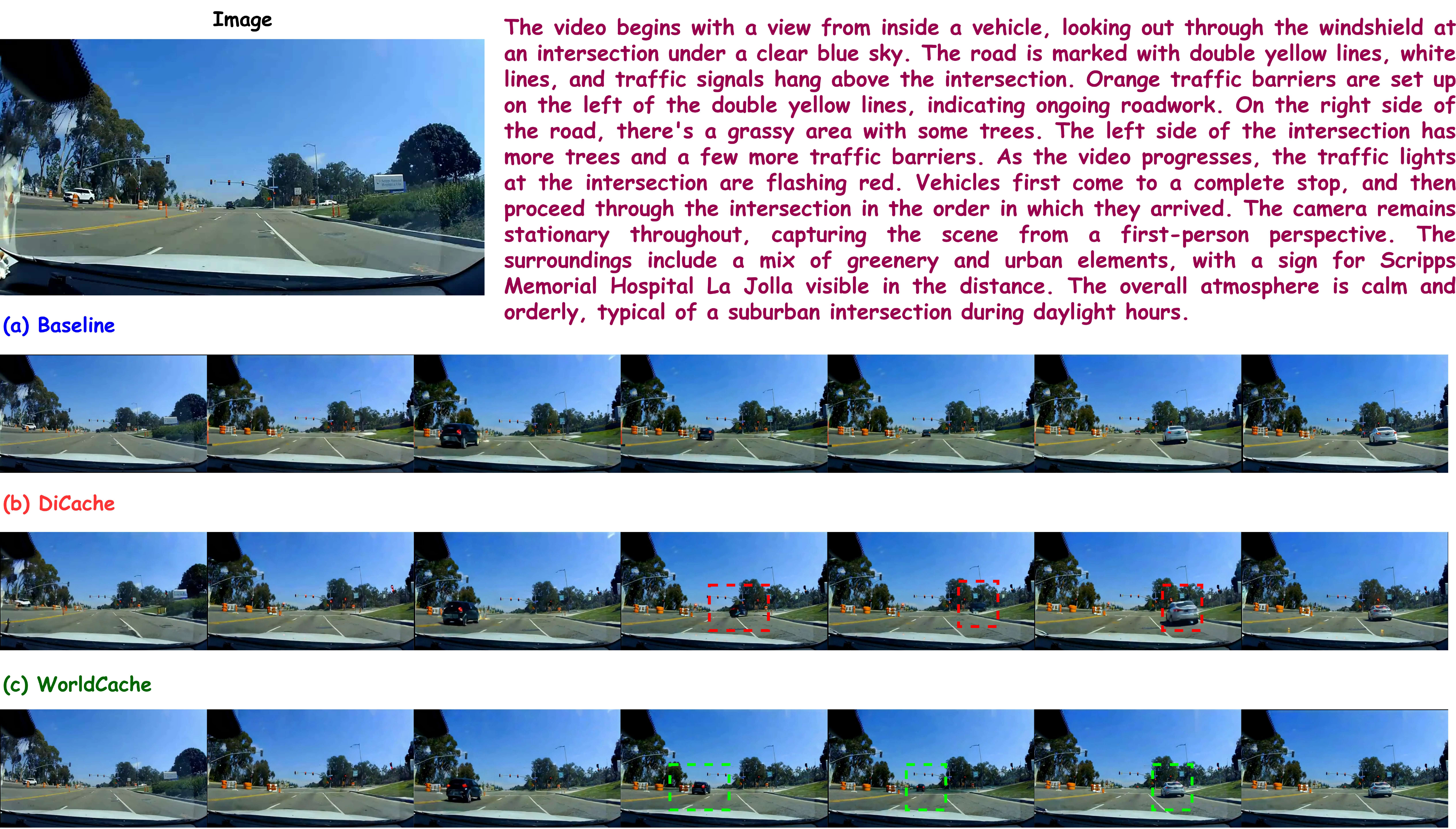}
    \caption{\textbf{Qualitative Image2World comparison (PAI-Bench).} Given the same conditioning image and text description, we show evenly-spaced frames from the generated rollout for (a) Baseline, (b) DiCache, and (c) WorldCache. DiCache exhibits temporal artifacts under dynamic scene evolution, e.g., object ghosting/deformation and inconsistent motion of vehicle and foreground structures (highlighted with \textcolor{red}{red dashed boxes}). WorldCache preserves the scene layout and maintains more coherent object appearance and trajectories over time (highlighted with \textcolor{green}{green dashed boxes}), while achieving substantial inference acceleration. Better viewed zoomed in.}
    \label{fig:qualitative-1}
    \vspace{-0.2in}
\end{figure}

\noindent \textbf{\textit{Qualitative Results.}} Fig.~\ref{fig:qualitative-1} demonstrates that DiCache introduces temporal artifacts (object ghosting, inconsistent motion) in dynamic scenes, highlighted by red dashed boxes, whereas WorldCache maintains coherent object appearance and consistent trajectories across frames (green dashed boxes). This qualitative comparison on PAI-Bench shows that WorldCache achieves both better temporal consistency and inference acceleration over the baseline and DiCache. More qualitative examples are presented in the supplementary material. 
Overall, we observe gains in the speed--quality boundary across both T2W and I2W and at both 2B and 14B scales. The WAN2.1 results further indicate that these benefits transfer across model families, supporting the view that perception and dynamics-constrained caching is a generally useful inference primitive for world-model generation. Overall, our findings point to a practical recipe for accelerating world-model rollouts: invest in decision/approximation quality early, then exploit the refinement phase to harvest large speedups with minimal impact on temporal coherence.

\subsection{Ablation Study}
\label{sec:ablations}

To isolate the contribution of each WorldCache module, we conduct incremental ablations on the 2B model under the I2W setting (Table~\ref{tab:ablation}). Besides, we also provide further ablations related to the key components in supplementary.

\begin{table}[t]
\centering
\caption{\textbf{Incremental ablation on Cosmos-Predict2.5-2B (I2W).} Each row adds one module to the previous configuration. \textit{Legend:} CFC = Causal Feature Caching (motion-adaptive thresholding); SWD = Saliency-Weighted Drift; OFA = Optimal Feature Approximation (Online System Identification); ATS = Adaptive Threshold Scheduling (dynamic threshold decay).}
\label{tab:ablation}
\vspace{4pt}
\resizebox{\textwidth}{!}{%
\begin{tabular}{l ccc cc}
\toprule
\textbf{Configuration} & \textbf{Domain} $\uparrow$ & \textbf{Quality} $\uparrow$ & \textbf{Overall} $\uparrow$ & \textbf{Speedup} $\uparrow$ & \textbf{Latency (s)} $\downarrow$ \\
\midrule
Base                          & 0.8447 & 0.7607 & 0.8027 & 1.00$\times$ & 55 \\
+ CFC                             & \textbf{0.8457} & 0.7583 & 0.8020 & 1.52$\times$ & 36 \\
+ CFC + SWD                       & 0.8414 & 0.7592 & 0.8003 & 1.67$\times$ & 33 \\
+ CFC + SWD + OFA                 & 0.8468 & \textbf{0.7602} & \textbf{0.8035} & 1.49$\times$ & 37 \\
+ CFC + SWD + OFA + ATS (\textbf{WorldCache}) & 0.8395 & 0.7559 & 0.7977 & \textbf{2.3}$\times$ & \textbf{25} \\
\bottomrule
\end{tabular}%
\vspace{-0.3in}
}

\end{table}

\noindent\textbf{CFC improves safety under motion.}
Adding CFC yields a 1.52$\times$ speedup (55$\rightarrow$36\,s) with essentially unchanged overall score (0.8020 vs.\ 0.8027). This indicates that motion-adaptive thresholding is an effective first-order control signal: it preserves fidelity by tightening reuse during fast dynamics, while still enabling frequent cache hits in stable intervals.

\noindent\textbf{SWD increases cache hits by focusing on salient regions.}
With SWD, speed increases further to 1.67$\times$ (33\,s). Although the global scores remain close to baseline, SWD improves decision quality by preventing background drift from dominating the skip criterion, thereby yielding additional cache hits without sacrificing foreground stability.

\noindent\textbf{OFA “invests” in approximation quality.}
Introducing OFA yields the highest-fidelity configuration (Overall 0.8035), slightly exceeding the baseline. The improved approximation, least-squares optimal blending (and motion-aligned reuse when enabled), reduces error accumulation on cache hits. This comes with added overhead and a more conservative hit pattern, which explains the reduced net speedup (1.49$\times$, 37\,s). In other words, OFA intentionally trades some throughput to raise the quality margin.

\noindent\textbf{ATS “spends” the quality margin for speed.}
Finally, ATS unlocks the largest acceleration by relaxing thresholds late in denoising, where refinement updates are small. With the stabilizing effects of CFC+SWD+OFA in place, ATS increases cache hits in the low-noise phase, improving speed to \textbf{2.30$\times$} (25\,s) while keeping overall within 0.6\% of baseline (0.7977 vs.\ 0.8027), matching the intended invest-and-spend effect.

\section{Conclusion}
\label{sec:conclusion}

We presented \textbf{WorldCache}, a unified framework for \emph{perception-constrained dynamical caching} in DiT-based video generation. By identifying the Zero-Order Hold assumption as the core source of artifacts in prior diffusion caching methods, we redesigned caching around motion-aware decisions (CFC), saliency-aligned drift estimation (SWD), improved approximation operators (OFA), and denoising-phase scheduling (ATS). WorldCache requires no training and no architectural changes, making it immediately deployable for accelerating next-generation video world models.

\bibliographystyle{splncs04}
\bibliography{main}






\clearpage
\appendix

\begin{center}
    {\Large \bfseries WorldCache: Content-Aware Caching for Accelerated Video World Models\par}
    \vspace{0.4em}
    {\large Supplementary Material\par}
\end{center}

\section*{Overview}
This supplementary material provides the following additional details:

\begin{itemize}
    \item \textbf{Evaluation Metrics:} Efficiency measures (latency/speedup), PAI-Bench-G quality protocol, and EgoDex-Eval fidelity metrics (PSNR/SSIM/LPIPS).
    \item \textbf{Implementation Details \& Runtime:} Training-free inference setup, evaluated backbones, hardware, and benchmark runtime considerations.
    \item \textbf{Baseline Comparisons:} Extended comparisons on PAI-Bench beyond the main paper.
    \item \textbf{Robotic Manipulation Results:} WorldCache evaluation on EgoDex-Eval.
    \item \textbf{Extended Technical Details:} Motion-compensated feature warping (OFA) and adaptive threshold scheduling (ATS).
    \item \textbf{Additional Ablations:} Hyperparameter analysis and denoising step budget effects.
    \item \textbf{Qualitative Results:} Caching failure modes and WorldCache mitigation examples.
    \item \textbf{Limitations \& Future Work:} Discuss some limitations and the potential future directions.
    
\end{itemize}

\section{Evaluation Metrics}
\label{sec:supp_metrics}

We report two complementary aspects of performance: \textbf{(i) quality} (how faithful and coherent the generated world rollout is) and \textbf{(ii) efficiency} (how fast the rollout can be generated). We evaluate quality on PAI-Bench using Domain and Quality scores with fine-grained sub-metrics, and additionally evaluate reconstruction-style fidelity on EgoDex-Eval using PSNR/SSIM/LPIPS. Efficiency is reported consistently across benchmarks using wall-clock latency and relative speedup for world generation. 

\subsection{Efficiency Metrics}
\label{sec:eff_metrics}

\paragraph{Latency.}
We report \textbf{Latency} as the end-to-end wall-clock time (in seconds) to generate one video sample under a fixed sampling configuration (same number of denoising steps, same resolution, same number of frames, same hardware and batch size). Latency includes the complete inference pipeline (probe computation, cache decision logic, approximation overhead, and any motion-compensation steps when enabled).

\paragraph{Speedup.}
We report \textbf{Speedup} relative to the unaccelerated baseline:
\begin{equation}
\text{Speedup} \;=\; \frac{\text{Latency}(\text{Baseline})}{\text{Latency}(\text{Method})}
\end{equation}
Higher speedup indicates faster generation.

\subsection{PAI-Bench Quality Metrics}
\label{sec:pai_metrics}

PAI-Bench~\cite{zhou2025pai} provides a multi-track evaluation framework for Physical AI.
In this work we focus on the generation track (PAI-Bench-G), which measures a world model's ability to synthesize coherent future videos under both Text-to-World (T2W) and Image-to-World (I2W) conditioning.
Following the official protocol, we evaluate on the full PAI-Bench-G suite comprising 1,044 samples and report both quality and efficiency metrics.
Quality is decomposed into a \emph{Quality Score} (eight perceptual dimensions) and a \emph{Domain Score} (seven semantic/physical categories); the two are averaged into an \textbf{Overall} score.

\subsubsection{Quality Score}
The Quality Score follows the VBench/VBench++ evaluation suite~\cite{huang2024vbench,huang2025vbenchplusplus} and comprises eight sub-metrics.
Throughout, we denote a generated video as a sequence of $T$ frames $\{f_1, f_2, \ldots, f_T\}$.

\paragraph{Subject Consistency (SC).}
This metric evaluates identity stability of the primary subject across frames.
We extract per-frame DINO~\cite{dino} features $\mathbf{d}_i$ (unit-normalised) and compute:
\begin{equation}
  \text{SC}
  \;=\;
  \frac{1}{T-1}\sum_{t=2}^{T}
  \frac{1}{2}\!\Big(
    \langle\mathbf{d}_1,\,\mathbf{d}_t\rangle
    +
    \langle\mathbf{d}_{t-1},\,\mathbf{d}_t\rangle
  \Big),
  \label{eq:sc}
\end{equation}
where $\langle\cdot,\cdot\rangle$ denotes cosine similarity.
The first term captures long-range consistency with the initial frame, while the second captures local (frame-to-frame) stability.

\paragraph{Background Consistency (BC).}
Background stability is assessed analogously, using CLIP~\cite{clip} image features $\mathbf{c}_i$ instead:
\begin{equation}
  \text{BC}
  \;=\;
  \frac{1}{T-1}\sum_{t=2}^{T}
  \frac{1}{2}\!\Big(
    \langle\mathbf{c}_1,\,\mathbf{c}_t\rangle
    +
    \langle\mathbf{c}_{t-1},\,\mathbf{c}_t\rangle
  \Big).
  \label{eq:bc}
\end{equation}

\paragraph{Motion Smoothness (MS).}
Motion plausibility is quantified via a frame-interpolation consistency test.
The video is subsampled by dropping all odd-indexed frames, which are then reconstructed using a pre-trained frame interpolation model~\cite{amt}.
Let $\hat{f}_{2k-1}$ denote the reconstructed version of the original frame $f_{2k-1}$.
The raw error is:
\begin{equation}
  S_{\text{smooth}}
  \;=\;
  \frac{1}{\lfloor T/2\rfloor}
  \sum_{k=1}^{\lfloor T/2\rfloor}
  \bigl\|f_{2k-1} - \hat{f}_{2k-1}\bigr\|_1,
  \label{eq:ms_raw}
\end{equation}
which is normalised to $[0,1]$ and inverted so that higher values indicate smoother motion:
\begin{equation}
  \text{MS}
  \;=\;
  1 - \frac{S_{\text{smooth}}}{255}.
  \label{eq:ms}
\end{equation}

\paragraph{Aesthetic Quality (AQ).}
Visual appeal, encompassing composition, colour harmony, and artistic quality, is scored per frame using the LAION aesthetic predictor~\cite{laion-ai} on a $[0,10]$ scale.
Scores are linearly mapped to $[0,1]$ and averaged:
\begin{equation}
  \text{AQ}
  \;=\;
  \frac{1}{T}\sum_{t=1}^{T}
  \frac{\text{LAION}(f_t)}{10}.
  \label{eq:aq}
\end{equation}

\paragraph{Imaging Quality (IQ).}
Low-level fidelity (noise, blur, exposure artefacts) is evaluated with the MUSIQ predictor~\cite{musiq}, yielding per-frame scores in $[0,100]$.
The video-level metric is:
\begin{equation}
  \text{IQ}
  \;=\;
  \frac{1}{T}\sum_{t=1}^{T}
  \frac{\text{MUSIQ}(f_t)}{100}.
  \label{eq:iq}
\end{equation}

\paragraph{Overall Consistency (OC).}
Semantic alignment between the generated video and the textual prompt is measured using the ViCLIP~\cite{internvid} video--text similarity score:
\begin{equation}
  \text{OC}
  \;=\;
  \text{ViCLIP}(\{f_1,\ldots,f_T\},\;\text{prompt}).
  \label{eq:oc}
\end{equation}

\paragraph{I2V Subject (IS).}
For image-conditioned generation, subject fidelity is measured by comparing the conditioning image $f_{\text{ref}}$ to generated frames using DINO features $\mathbf{s}_i$:
\begin{equation}
  \text{IS}
  \;=\;
  \frac{1}{T-1}\sum_{t=2}^{T}
  \frac{1}{2}\!\Big(
    \langle\mathbf{s}_{\text{ref}},\,\mathbf{s}_t\rangle
    +
    \langle\mathbf{s}_{t-1},\,\mathbf{s}_t\rangle
  \Big).
  \label{eq:is}
\end{equation}

\paragraph{I2V Background (IB).}
Background and layout fidelity in image-conditioned generation is computed using DreamSim~\cite{dreamsim} features $\mathbf{b}_i$, which are sensitive to spatial layout:
\begin{equation}
  \text{IB}
  \;=\;
  \frac{1}{T-1}\sum_{t=2}^{T}
  \frac{1}{2}\!\Big(
    \langle\mathbf{b}_{\text{ref}},\,\mathbf{b}_t\rangle
    +
    \langle\mathbf{b}_{t-1},\,\mathbf{b}_t\rangle
  \Big).
  \label{eq:ib}
\end{equation}

\paragraph{Quality aggregation.}
The Quality Avg is computed as the arithmetic mean of the eight sub-metrics:
\begin{equation}
  \text{Quality Avg}
  \;=\;
  \frac{1}{8}
  \sum_{m\,\in\,\{\text{SC,\,BC,\,MS,\,AQ,\,IQ,\,OC,\,IS,\,IB}\}} m.
  \label{eq:quality_avg}
\end{equation}

\subsubsection{Domain Score (Semantic/Physical Plausibility)}
\label{sec:domain_score}

While Quality Score captures perceptual fidelity, it does not assess whether generated dynamics are \emph{physically plausible}.
Domain Score addresses this gap through an \emph{MLLM-as-Judge} protocol:
a strong vision-language model, specifically Qwen3-VL-235B-A22B-Instruct~\cite{bai2025qwen3}, is queried with uniformly sampled frames and a curated set of $Q$ binary verification questions $\{q_j\}_{j=1}^{Q}$ that encode expected physical and semantic constraints (\eg, ``Does the robotic arm lift the object?'').
For each question the judge emits a binary response $\hat{a}_j \in \{\texttt{YES},\texttt{NO}\}$.
The Domain Score is then the accuracy of these responses against the ground-truth labels $a_j$:
\begin{equation}
  \text{Domain}
  \;=\;
  \frac{1}{Q}\sum_{j=1}^{Q}
  \mathbf{1}\!\bigl[\hat{a}_j = a_j\bigr].
  \label{eq:domain}
\end{equation}
Questions are organised into seven semantic categories reflecting distinct facets of world-model reasoning:
\textbf{CS} (Common Sense),
\textbf{AV} (Autonomous Vehicle),
\textbf{RO} (Robot),
\textbf{IN} (Industry),
\textbf{HU} (Human),
\textbf{PH} (Physics), and
\textbf{MI} (Miscellaneous).
We report per-category accuracy and compute \textbf{Domain Avg} as their mean:
\begin{equation}
  \text{Domain Avg}
  \;=\;
  \frac{1}{7}
  \sum_{c\,\in\,\{\text{CS,\,AV,\,RO,\,IN,\,HU,\,PH,\,MI}\}} c.
  \label{eq:domain_avg}
\end{equation}

\subsubsection{Overall Score}
\label{sec:overall_score}

Following the PAI-Bench convention, we summarise generation quality with a single scalar that equally weights perceptual fidelity and physical plausibility:
\begin{equation}
  \text{Overall}
  \;=\;
  \frac{1}{2}
  \bigl(\text{Domain Avg} + \text{Quality Avg}\bigr).
  \label{eq:overall}
\end{equation}

\subsection{EgoDex-Eval Fidelity Metrics}
\label{sec:egodex_metrics}

To complement PAI-Bench with reconstruction-style fidelity measures, we also evaluate on EgoDex-Eval~\cite{hoque2505egodex}, reporting standard full-reference image/video metrics alongside efficiency.

\paragraph{PSNR.}
Peak Signal-to-Noise Ratio measures pixel-level reconstruction fidelity between generated and reference frames. Higher PSNR indicates lower distortion.

\paragraph{SSIM.}
Structural Similarity Index measures perceived structural similarity (luminance, contrast, and structure) between generated and reference frames. Higher SSIM indicates better structural preservation.

\paragraph{LPIPS.}
Learned Perceptual Image Patch Similarity measures perceptual distance using deep features. Lower LPIPS indicates closer perceptual similarity to the reference.

\section{Implementation Details and Runtime Setup}
\label{sec:supp_impl_setup}

\paragraph{Training-free inference acceleration.}
WorldCache is a strictly \textbf{training-free} and \textbf{plug-and-play} method. It does not modify model weights and can be enabled/disabled at runtime. All improvements are obtained by (i) deciding when to reuse cached intermediate activations and (ii) applying a lightweight cache-hit approximation, both of which can be toggled on/off at runtime.

\paragraph{Evaluated backbones.}
\textbf{Cosmos-Predict2.5}~\cite{ali2025world} is a diffusion transformer (DiT) based world model that supports both \emph{Text2World} and \emph{Image2World} generation within a unified sampling pipeline. \textbf{WAN2.1}~\cite{wan2025wan} is an open large-scale diffusion-transformer video model suite, released with multiple parameter scales (e.g., 1.3B/14B). \textbf{DreamDojo}~\cite{gao2026dreamdojo} is a robot-oriented interactive world model trained from large-scale human egocentric videos, released with pretrained/post-trained checkpoints (e.g., 2B/14B) and evaluation sets. In our EgoDex-Eval~\cite{hoque2505egodex} experiments, we evaluate WorldCache on the provided model checkpoints without modifying training.

\paragraph{Hardware and codebases.}
All experiments are run on a \textbf{single NVIDIA H200 (140\,GB)} GPU, except for the calculation of the Domain score in PAI-bench, as that uses 4 NVIDIA H200 140 GB GPUs because of Qwen3-VL235B-A22B-Instruct~\cite{bai2025qwen3} used as a judge. We use the \emph{official} inference codebases and released checkpoints for all evaluated backbones: Cosmos-Predict2.5, WAN2.1, and DreamDojo. Unless otherwise specified, we keep the generation configuration fixed across methods within each backbone (same resolution, video length, denoising steps, scheduler, guidance setting, and batch size). Reported numbers are reproduced from our runs under the same hardware and software environment.

All compared caching baselines are run under identical generation settings and hardware. Hyperparameters for WorldCache are selected on a held-out subset and then fixed for the full benchmark. No method uses additional training data or model updates during evaluation.

\paragraph{PAI-Bench-G evaluation (world-model generation).}
For world-model generation quality, we follow the \textbf{PAI-Bench-G} track \cite{zhou2025pai} and evaluate on the full suite of \textbf{1,044 samples} under both T2W and I2W. We report the benchmark-defined quality metrics (Domain/Quality/Overall as described in Sec.~\ref{sec:pai_metrics}) and efficiency metrics (Latency/Speedup). Latency is measured end-to-end from the start of the denoising loop to the completion of decoding, and includes probe computation, cache decision logic, and any approximation overhead (e.g., OFA warping). To highlight evaluation-scale impact, we also report the total runtime to process the entire benchmark:
\begin{equation}
T_{\text{PAI}} \;=\; 1044 \times T_{\text{avg}},
\end{equation}
where $T_{\text{avg}}$ is the mean per-sample latency in our tables. For example, on Cosmos-2B (I2W), the baseline runtime corresponds to $55.04\times1044\approx 16$\,hours, whereas WorldCache at 24.48\,s/sample corresponds to $\approx 7.1$\,hours, saving $\approx 9$\,hours per full PAI-Bench-G run on the same hardware.

\paragraph{EgoDex-Eval evaluation (ground-truth-conditioned robotics video).}
To evaluate WorldCache on downstream robotics prediction with ground-truth video, we use EgoDex-Eval, an egocentric manipulation benchmark derived from the EgoDex dataset~\cite{hoque2505egodex}. Following the standard protocol used in robot world-model evaluation, we condition on the first frame and generate a rollout (e.g., 81 frames in our WAN2.1-14B setup), then compute frame-level full-reference metrics\cite{khan2024guardian}: PSNR$\uparrow$, SSIM$\uparrow$, and LPIPS$\downarrow$, together with Latency/Speedup. As with PAI-Bench, we measure end-to-end wall-clock latency for the full generation pipeline. When reporting evaluation-scale time, we use:
\begin{equation}
T_{\text{EgoDex}} \;=\; N_{\text{eval}} \times T_{\text{avg}},
\end{equation}
where $N_{\text{eval}}$ is the number of evaluation episodes and $T_{\text{avg}}$ is the mean per-episode latency.

\section{More Baseline Comparison}
\label{sec:more_baselines}

To further contextualize WorldCache against recent training-free caching baselines, we compare with EasyCache~\cite{zhou2025less} and TeaCache~\cite{liu2025timestep} under the same Cosmos-Predict2.5-2B setup for both Image2World (I2W) and Text2World (T2W) on PAI-Bench (Table~\ref{tab:easycache_teacache}). We additionally include DiCache as a strong probe-based caching baseline. In Table~\ref{tab:easycache_teacache}, \textbf{bold}/\underline{underline} indicate the best/second-best results among accelerated methods only (excluding the unaccelerated baseline).

\noindent \textbf{Text2World (T2W).}
For T2W, TeaCache (Slow) achieves the best accelerated Overall (0.7454) and the second-best Quality (0.7274), but again is conservative (49.40\,s, 1.1$\times$). EasyCache yields the best accelerated Domain (0.7641), while DiCache provides the strongest speed among non-WorldCache baselines (40.82\,s, 1.3$\times$). WorldCache reduces latency to 26.28\,s (\textbf{2.10$\times$}) with Overall 0.7450, matching the quality of the strongest baseline variants while delivering substantially higher acceleration.

\noindent \textbf{Image2World (I2W).}
Among prior caching baselines, EasyCache attains the best Domain score (0.8399), while TeaCache (Slow) achieves the strongest quality and overall among non-WorldCache methods (Quality 0.7562, Overall 0.7979) but at conservative speed (49.59\,s, 1.1$\times$). DiCache and TeaCache (Fast) are faster ($\sim$40--41\,s, 1.3--1.3$\times$) but with lower overall scores (0.7941--0.7965). WorldCache provides a substantially better efficiency point with \textbf{2.30$\times$} speedup (55.04\,s $\rightarrow$ 24.48\,s) while keeping overall competitive (0.7977), and close to the best accelerated overall score and far faster than all other caching methods.

Overall, existing caching baselines cluster around $\sim$1.3$\times$ speedup (or trade speed for slightly higher scores), whereas WorldCache consistently exceeds more than $2\times$ speedup while remaining within the same quality band on both I2W and T2W.

\begin{table}[t!]
\centering
\caption{\textbf{PAI-Bench comparison with EasyCache and TeaCache (Cosmos-2B).} We report Domain, Quality, Overall, latency (s), and speedup vs.\ baseline for both Image2World (I2W) and Text2World (T2W). \textbf{Bold} and \underline{underline} denote the best and second-best \emph{among accelerated methods only} (excluding Baseline) within each block.}
\label{tab:easycache_teacache}
\vspace{2pt}
\setlength{\tabcolsep}{9pt}
\renewcommand{\arraystretch}{1.08}
\small
\resizebox{\columnwidth}{!}{%
\begin{tabular}{l c c c c c}
\toprule
\textbf{Method} & \textbf{Domain} $\uparrow$ & \textbf{Quality} $\uparrow$ & \textbf{Overall} $\uparrow$ & \textbf{Lat.\,(s)} $\downarrow$ & \textbf{Speedup} $\uparrow$ \\
\midrule
\rowcolor{green!10}
\multicolumn{6}{c}{\textbf{Text2World (T2W) --- Cosmos-Predict2.5-2B}} \\
\midrule
Baseline            & 0.7670 & 0.7280 & 0.7475 & 54.34 & 1.0$\times$ \\
\midrule
EasyCache           & \textbf{0.7641} & 0.7262 & \underline{0.7451} & 41.41 & 1.3$\times$ \\
DiCache             & 0.7590 & 0.7272 & 0.7431 & \underline{40.82} & \underline{1.3$\times$} \\
TeaCache (Fast)     & 0.7616 & 0.7266 & 0.7448 & 41.07 & 1.4$\times$ \\
TeaCache (Slow)     & 0.7634 & \underline{0.7274} & \textbf{0.7454} & 49.40 & 1.1$\times$ \\
\midrule
WorldCache (Ours)   & 0.7630 & 0.7270 & 0.7450 & \textbf{26.28} & \textbf{2.10$\times$} \\
\midrule
\rowcolor{orange!10}
\multicolumn{6}{c}{\textbf{Image2World (I2W) --- Cosmos-Predict2.5-2B}} \\
\midrule
Baseline            & 0.8450 & 0.7610 & 0.8030 & 55.04 & 1.0$\times$ \\
\midrule
EasyCache           & \textbf{0.8399} & 0.7552 & 0.7975 & 40.25 & 1.3$\times$ \\
DiCache             & 0.8352 & 0.7522 & 0.7941 & \underline{39.68} & \underline{1.3$\times$} \\
TeaCache (Fast)     & 0.8381 & 0.7549 & 0.7965 & 41.00 & 1.3$\times$ \\
TeaCache (Slow)     & \underline{0.8396} & \textbf{0.7562} & \textbf{0.7979} & 49.59 & 1.1$\times$ \\
\midrule
WorldCache (Ours)   & 0.8395 & \underline{0.7559} & \underline{0.7977} & \textbf{24.48} & \textbf{2.30$\times$} \\
\bottomrule
\end{tabular}%
}
\end{table}

\section{Evaluation Results on Robotic Manipulation: EgoDex-Eval}
\label{sec:egodex}

To evaluate WorldCache on a downstream robotics setting with ground-truth video supervision, we benchmark on EgoDex-Eval~\cite{hoque2505egodex}, an egocentric robot manipulation dataset. We condition each model on the first frame of an episode and generate rollouts, reporting frame-level PSNR and SSIM ($\uparrow$) and LPIPS ($\downarrow$) against the ground-truth video, along with end-to-end latency and speedup (Table~\ref{tab:egodex_all}).

\begin{table}[t]
\centering
\caption{\textbf{EgoDex-Eval results (I2V) across backbones.} We report frame-level PSNR/SSIM/LPIPS against ground-truth videos, along with end-to-end latency and speedup relative to the unaccelerated baseline. \textbf{Bold} denotes the best value per block (including Baseline), and \underline{underline} denotes the best \emph{accelerated} result (DiCache/WorldCache).}
\label{tab:egodex_all}
\vspace{2pt}
\setlength{\tabcolsep}{12.0pt}
\renewcommand{\arraystretch}{1.08}
\small
\resizebox{\textwidth}{!}{%
\begin{tabular}{l c c c c c}
\toprule
\textbf{Method}
& \textbf{PSNR} $\uparrow$ & \textbf{SSIM} $\uparrow$ & \textbf{LPIPS} $\downarrow$
& \textbf{Speedup} $\uparrow$ & \textbf{Lat.\,(s)} $\downarrow$ \\
\midrule

\rowcolor{blue!6}
\multicolumn{6}{l}{\textit{WAN2.1-14B (I2V)}} \\
Baseline   & \textbf{13.30} & \textbf{0.503} & \textbf{0.459} & 1.00$\times$ & 391.9 \\
DiCache    & 12.95 & 0.491 & 0.461 & 1.88$\times$ & 208.6 \\
WorldCache & \underline{13.19} & \underline{0.498} & \underline{0.460} & \underline{\textbf{2.30$\times$}} & \underline{\textbf{171.6}} \\
\midrule

\rowcolor{green!6}
\multicolumn{6}{l}{\textit{Cosmos-Predict-2.5-2B (I2V)}} \\
Baseline   & \textbf{12.87} & \underline{0.455} & \textbf{0.518} & 1.00$\times$ & 70.01 \\
DiCache    & 12.63 & 0.445 & 0.531 & 1.34$\times$ & 51.97 \\
WorldCache & \underline{12.82} & \textbf{0.466} & \underline{\textbf{0.518}} & \underline{\textbf{1.62$\times$}} & \underline{\textbf{43.24}} \\
\midrule

\rowcolor{orange!8}
\multicolumn{6}{l}{\textit{DreamDojo-2B (I2V)}} \\
Baseline   & 23.63 & \textbf{0.775} & \textbf{0.226} & 1.00$\times$ & 19.73 \\
DiCache    & 20.41 & 0.734 & 0.252 & 1.58$\times$ & 12.46 \\
WorldCache & \underline{\textbf{23.69}} & \underline{0.737} & \underline{0.251} & \underline{\textbf{1.90$\times$}} & \underline{\textbf{10.36}} \\
\bottomrule
\end{tabular}%
}
\end{table}

\noindent \textbf{WAN2.1-14B \cite{wan2025wan} (I2V).}
WorldCache achieves a \textbf{2.30$\times$} speedup (391.9\,s $\rightarrow$ 171.6\,s) while remaining close to baseline quality: PSNR drops marginally (13.19 vs.\ 13.30; $\sim$99.2\% retention), SSIM remains high (0.498 vs.\ 0.503), and LPIPS is nearly unchanged (0.460 vs.\ 0.459). DiCache is also faster than baseline (1.88$\times$, 208.6\,s) but exhibits a larger fidelity gap across all three metrics (12.95 PSNR, 0.491 SSIM, 0.461 LPIPS). This setting is particularly challenging for caching because egocentric manipulation contains continuous hand--object contact and fine-grained motion, where stale reuse can accumulate errors. The results indicate WorldCache better preserves motion and appearance while accelerating inference.

\noindent \textbf{Cosmos-2.5-2B \cite{ali2025world} (I2V).}
On Cosmos-Predict-2.5-2B, WorldCache improves both efficiency and fidelity relative to DiCache. It reaches \textbf{1.62$\times$} speedup (70.01\,s $\rightarrow$ 43.24\,s) while preserving PSNR (12.82 vs.\ 12.87) and matching the best LPIPS (0.518, equal to baseline). Notably, WorldCache also attains a higher SSIM (0.466) than both baseline (0.455) and DiCache (0.445), suggesting improved structural stability under caching. DiCache provides a smaller \textbf{1.34$\times$} speedup (51.97\,s) and shows larger degradation in PSNR and LPIPS (12.63 PSNR, 0.531 LPIPS). Overall, this block demonstrates that WorldCache’s caching strategy transfers beyond a single backbone under ground-truth-conditioned evaluation.

\noindent \textbf{DreamDojo-2B \cite{gao2026dreamdojo} (I2V).}
For DreamDojo-2B, WorldCache also delivers a favorable trade-off by achieving \textbf{1.90$\times$} speedup (19.73\,s $\rightarrow$ 10.36\,s) while preserving PSNR (23.69 vs.\ 23.63) with moderate changes in SSIM and LPIPS (0.737 vs.\ 0.775; 0.251 vs.\ 0.226). In contrast, DiCache yields a smaller \textbf{1.58$\times$} speedup (12.46\,s) and suffers a pronounced fidelity loss (PSNR 20.41, SSIM 0.734, LPIPS 0.252). Overall, across WAN2.1, Cosmos, and DreamDojo, WorldCache consistently outperforms DiCache, achieving higher speedups with lower quality degradation under EgoDex-Eval.

\section{Extended Technical Details}

\subsection{Motion-Compensated Feature Warping in OFA}
\label{sec:supp_flow_scaling}

In Section~\ref{sec:ofa}, we described the Optimal Feature Alignment (OFA) mechanism, which estimates a displacement field to warp cached features from the previous timestep to the current coordinate frame. Because computing dense optical flow directly on high-resolution, deep semantic feature maps can introduce significant computational overhead and susceptibility to high-frequency activation noise, we implement a spatial flow-scaling technique. Specifically, we introduce a spatial downsampling factor, $s_{\text{flow}}$, applied prior to the displacement calculation.

During inference, the latent features $\mathbf{z}_{t}^{(0)}$ and $\mathbf{z}_{t-1}^{(0)}$ are first downsampled to a spatial resolution of $s_{\text{flow}} \times H \times s_{\text{flow}} \times W$ via bilinear interpolation. The Lucas-Kanade optical flow \cite{bruhn2005lucas} equations are solved on this lower-resolution grid to produce a coarse displacement field. Finally, this continuous displacement field is upsampled back to the original resolution, and its vector magnitudes are scaled by $1 / s_{\text{flow}}$ to ensure correct geometric mapping during the final warping operation.

This design choice serves two critical functions. First, it acts as a strong spatial low-pass filter, forcing the flow estimation to focus on macroscopic structural motion rather than microscopic, high-frequency signal fluctuations inherent to deep transformer representations. Second, by computing the correlation matrix on feature maps that are $1/25$\textsuperscript{th} the spatial area, the runtime complexity of the Lucas-Kanade solver is drastically reduced, ensuring the alignment mechanism adds less than 3\% computational overhead to the caching pipeline.

\subsection{Adaptive Thresholding Scheduling (ATS)}
\label{sec:supp_ats_implementation}

In Section~3.7 of the main text, we introduced the Adaptive Threshold Scheduling (ATS) mechanism, which relaxes the caching threshold $\tau_{\text{ATS}}(t)$ as the generative process transitions from global structure formation to high-frequency detail refinement. In our practical implementation, this relaxation is controlled by a temporal decay factor applied dynamically at each step.

\textbf{Quadratic Threshold Decay.} 
While Equation~13 describes the conceptual linear relaxation governed by $\beta_d$, our exact implementation employs a quadratic scaling function to provide a smoother transition across the denoising trajectory. Let $N$ be the total number of diffusion steps (e.g., \texttt{worldcache\_num\_steps = 35} in our standard configuration), and let $t \in [0, N-1]$ be the current forward sampling step index. 
We define a normalized progress variable $u = N / 35.0$ to ensure the decay curve remains scale-invariant regardless of the user's chosen total number of steps. The base multiplier coefficient $C(u)$ is derived via a quadratic fit:
\begin{equation}
    C(u) = \frac{u^2}{6} + \frac{u}{2} + \frac{10}{3}.
\end{equation}
At step $t$, the step ratio $r_t = t / N$ is computed, and the final dynamic decay factor $D(t)$ applied to the base threshold is calculated as:
\begin{equation}
    D(t) = 1.0 + C(u) \cdot r_t.
    \label{eq:ats_decay_factor}
\end{equation}
Thus, the final threshold at step $t$ is strictly given by $\tau_{\text{ATS}}(t) = \tau_{\text{base}} \cdot D(t)$. 

\textbf{Boundary Behavior.}
This specific quadratic fit was empirically designed to hit key operational targets: it tightly bounds the threshold scaling near $1.0$ at $t=0$ (enforcing rigorous structural computation) and smoothly accelerates the relaxation multiplier to approximately $5.0$ as $t \to N$ (when $N=35$). This ensures that during the final $\sim 20\%$ of the generation process, where latent updates are minimal, the network aggressively reuses cached features, yielding maximum acceleration without degrading spatial fidelity.

\begin{table}[t!]
\centering
\caption{\textbf{Hyperparameter sensitivity for skip decisions (stage-wise).} Each block varies a single hyperparameter while keeping the rest fixed in the corresponding stage configuration: CFC is swept with only CFC enabled; SWD is swept with CFC fixed to $\alpha{=}2$; ATS is swept with CFC+SWD+OFA fixed to defaults. Results: Cosmos-Predict2.5--2B (I2W). \textbf{Bold} indicates the chosen default.}
\label{tab:hparam_decision}
\vspace{2pt}
\setlength{\tabcolsep}{15pt}
\renewcommand{\arraystretch}{1.00}
\small
\resizebox{0.9\textwidth}{!}{%
\begin{tabular}{l c c c c}
\toprule
\textbf{Hyperparameter} & \textbf{Value} & \textbf{Domain} $\uparrow$ & \textbf{Quality} $\uparrow$ & \textbf{Overall} $\uparrow$ \\
\midrule
\multirow{4}{*}{CFC: $\alpha$}
& 0.1   & 0.8389 & 0.7362 & 0.7876 \\
& \textbf{0.2}   & \textbf{0.8457} & \textbf{0.7583} & \textbf{0.8020} \\
& 0.4   & 0.8417 & 0.7374 & 0.7896 \\
& 0.5 & 0.8362 & 0.7356 & 0.7859 \\
\midrule
\multirow{4}{*}{SWD: $\beta_s$}
& 0.05 & 0.8483 & 0.7335 & 0.7909 \\
& \textbf{0.12} & \textbf{0.8414} & \textbf{0.7592} & \textbf{0.8003} \\
& 0.5  & 0.8420 & 0.7402 & 0.7911 \\
& 1    & 0.8390 & 0.7344 & 0.7867 \\
\midrule
\multirow{4}{*}{ATS: $\beta_d$}
& 2 & 0.8330 & 0.7486 & 0.7908 \\
& \textbf{4} & \textbf{0.8395} & \textbf{0.7559} & \textbf{0.7977} \\
& 6 & 0.8314 & 0.7483 & 0.7899 \\
& 8 & 0.8289 & 0.7392 & 0.7841 \\
\bottomrule
\end{tabular}%
}
\end{table}

\begin{table}[t!]
\centering
\caption{\textbf{Hyperparameter sensitivity for cache-hit approximation (stage-wise).} OFA operator and warp scale sweeps are run with CFC and SWD fixed to defaults. Results: Cosmos-Predict2.5--2B (I2W). \textbf{Bold} indicates the chosen default.}
\label{tab:hparam_approx}
\vspace{2pt}
\setlength{\tabcolsep}{15pt}
\renewcommand{\arraystretch}{1.00}
\small
\resizebox{0.9\textwidth}{!}{%
\begin{tabular}{l c c c c}
\toprule
\textbf{Hyperparameter} & \textbf{Value} & \textbf{Domain} $\uparrow$ & \textbf{Quality} $\uparrow$ & \textbf{Overall} $\uparrow$ \\
\midrule
\multirow{3}{*}{OFA operator}
& OSI only   & 0.8478 & 0.7360 & 0.7919 \\
& Warp only  & 0.8402 & 0.7297 & 0.7850 \\
& \textbf{OSI + Warp} & \textbf{0.8468} & \textbf{0.7602} & \textbf{0.8035} \\
\midrule
\multirow{3}{*}{Warp scale}
& 0.2 & 0.8385 & 0.7309 & 0.7847 \\
& \textbf{0.5} & \textbf{0.8495} & \textbf{0.7420} & \textbf{0.7958} \\
& 1.0 & 0.8408 & 0.7357 & 0.7883 \\
\bottomrule
\end{tabular}%
}
\end{table}

\begin{figure}[t!]
    \centering
    \includegraphics[width=\textwidth]{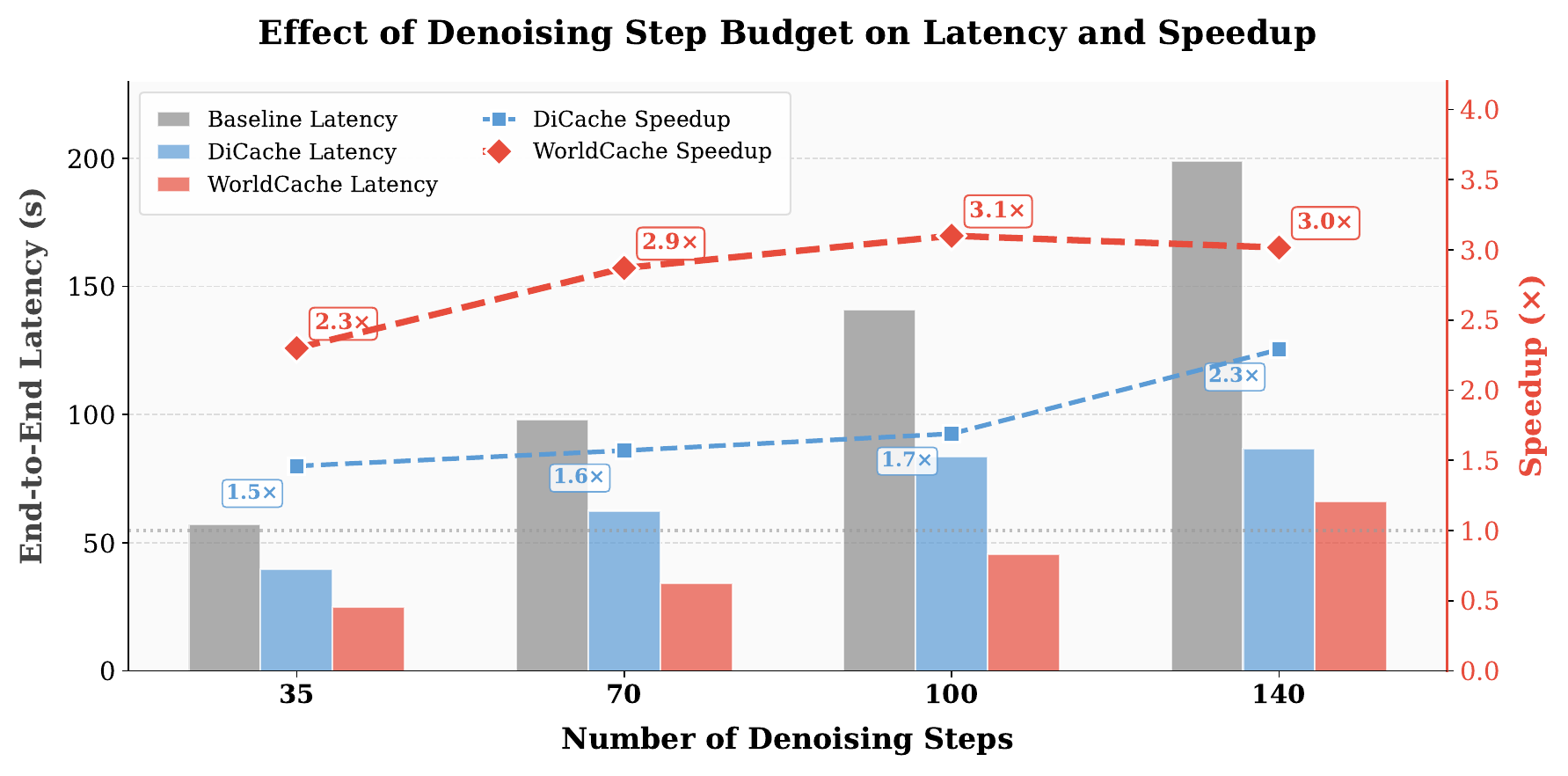}
    \caption{\textbf{Effect of the denoising step budget.} End-to-end latency and speedup of WorldCache vs. baselines when varying the number of denoising steps.}
    \label{fig:steps_ablation}
\end{figure}

\section{Additional Ablations}

\begin{figure}[t!]
    \centering
    \includegraphics[width=\textwidth]{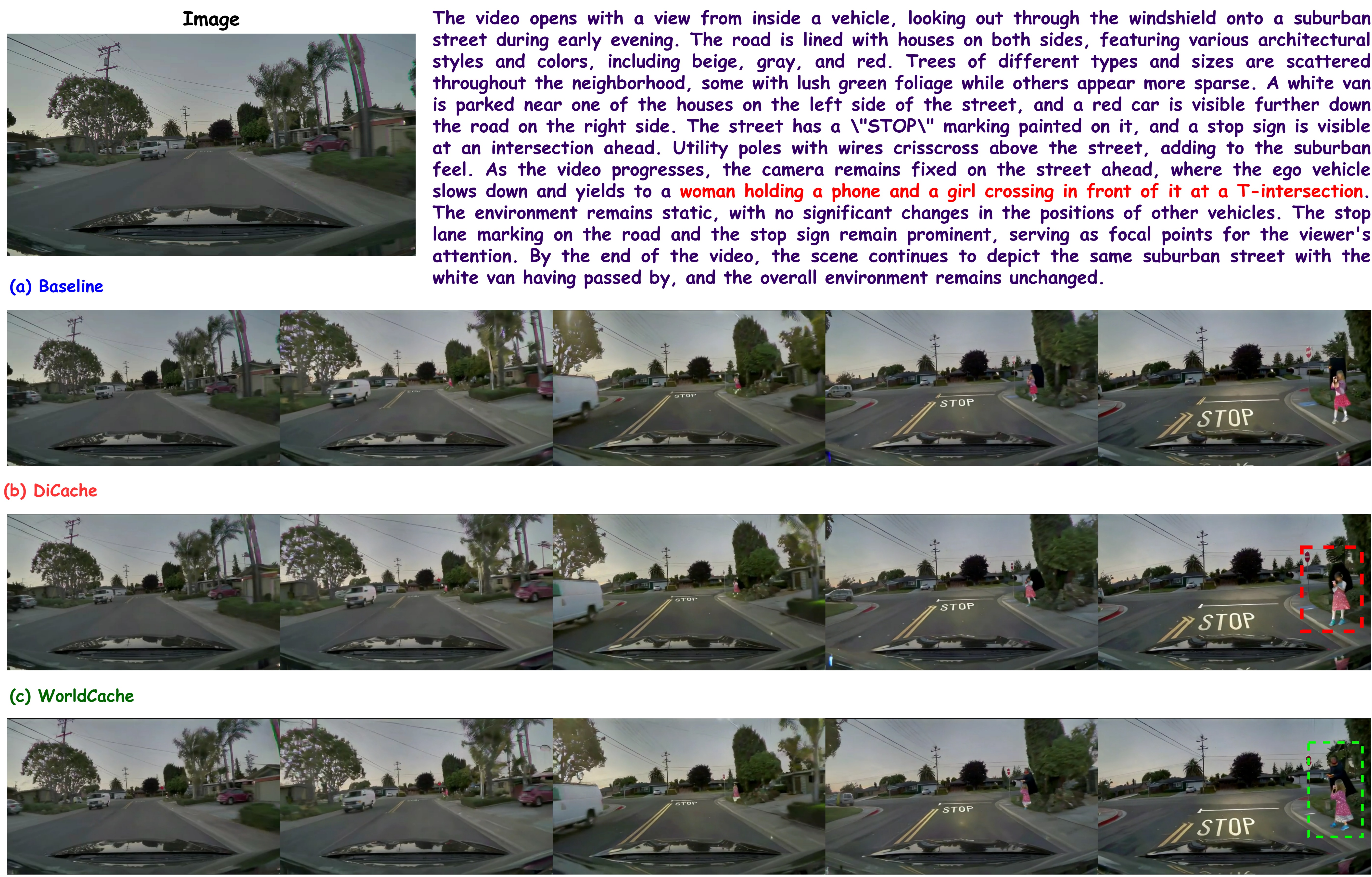}
    \caption{\textbf{Qualitative Image2World comparison on PAI-Bench (Cosmos-2B).} The conditioning image (top-left) and text description (top-right) specify a dashcam scene where the ego vehicle slows at a T-intersection as a woman and a child cross in front. We show evenly spaced frames from rollouts generated by (a) Baseline, (b) DiCache, and (c) WorldCache. DiCache exhibits temporal inconsistency on salient, moving entities, e.g., the pedestrians appearance and position become unstable and partially ghosted near the end of the rollout \textcolor{red}{(red dashed box)}. WorldCache maintains more coherent pedestrian identity and motion while preserving scene layout \textcolor{green}{(green dashed box)}, producing rollouts closer to the baseline under substantially lower inference latency. Better viewed zoomed in.}
    \label{fig:quality-1}
\end{figure}

\begin{figure}[t!]
    \centering
    \includegraphics[width=\textwidth]{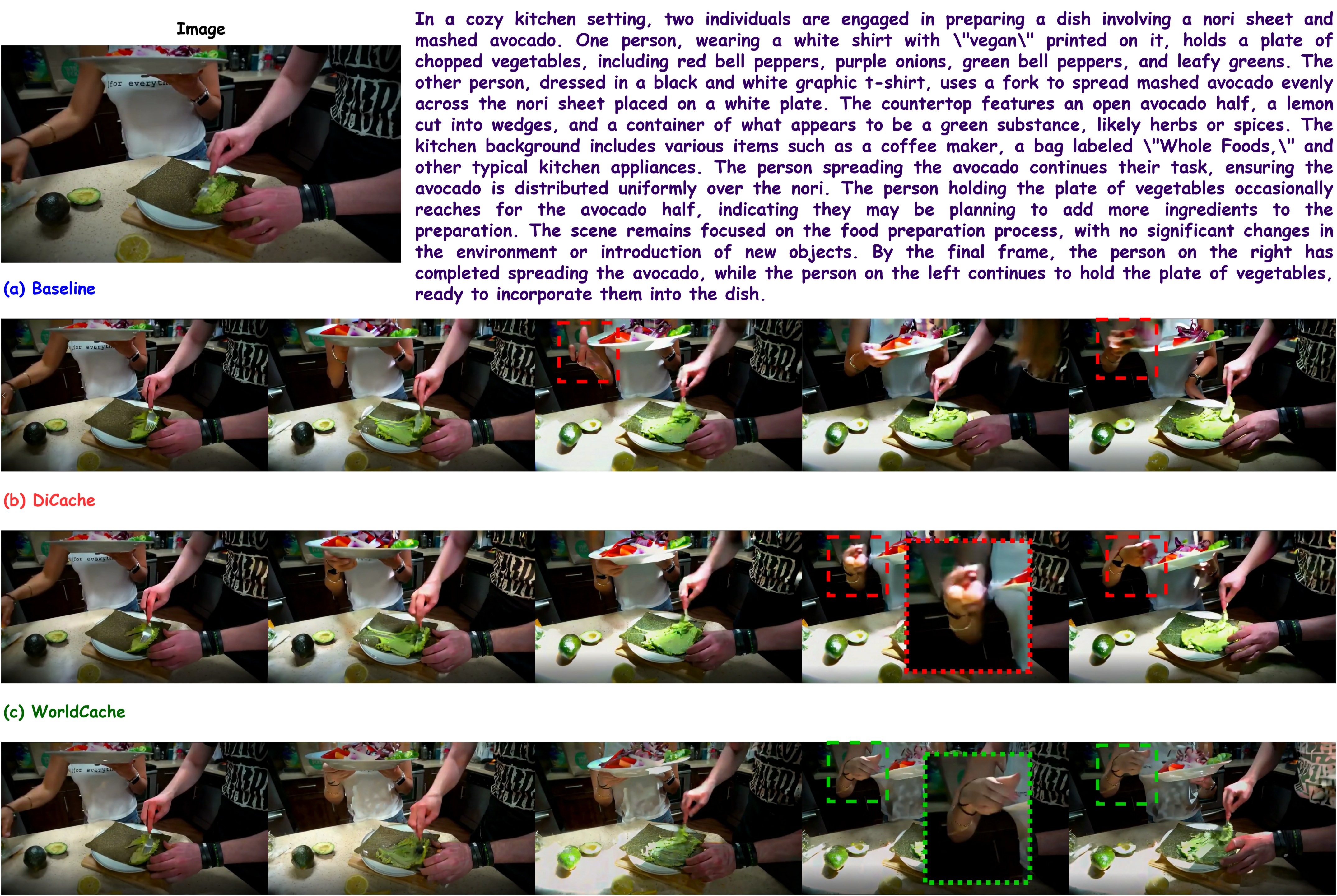}
    \caption{\textbf{Qualitative Image2World comparison on PAI-Bench (Cosmos-14B).}
    Given the conditioning image (top-left) and description (top-right), we generate a short kitchen rollout in which two people prepare food, where one spreads avocado on a nori sheet while the other holds a plate of vegetables. We visualize evenly spaced frames from (a) the unaccelerated baseline, (b) DiCache, and (c) WorldCache. While the baseline struggles with hand pose and object geometry, DiCache also exhibits noticeable temporal instability in the most salient, high-frequency regions, particularly around the hands and the plate: edges wobble, fine structures deform, and the appearance of the plate/hand region changes inconsistently over time (\textcolor{red}{red dashed boxes}). WorldCache largely avoids these artifacts, preserving sharper boundaries and more coherent local motion in the interaction region (\textcolor{green}{green dashed boxes}), yielding outputs that are closer to the baseline at substantially reduced latency. Best viewed with zoom.}
    \label{fig:quality-2}
\end{figure}

\subsection{Hyperparameter Selection}
\label{sec:supp_hparam}

To avoid ambiguity, we report hyperparameter studies as two separate ablations: (i) \textbf{skip-decision sensitivity} (Table~\ref{tab:hparam_decision}) and (ii) \textbf{cache-hit approximation sensitivity} (Table~\ref{tab:hparam_approx}). In all cases, rows within a block are not applied sequentially, as each row corresponds to a \textbf{separate run} where only the listed hyperparameter is changed, and all other settings are held fixed.

\paragraph{Skip-decision hyperparameters.}
Table~\ref{tab:hparam_decision} varies the parameters that control when caching is allowed. CFC benefits from increasing motion sensitivity up to $\alpha{=}2$, after which performance drops, so we select $\alpha{=}2$. SWD performs best at $\beta_s{=}0.12$, indicating that moderate saliency weighting improves skip decisions while overly strong weighting reduces robustness. For ATS, $\beta_d{=}4$ yields the best trade-off as larger values over-relax the reuse criterion and degrade both Domain and Quality, consistent with excessive skipping in late denoising.

\paragraph{Approximation hyperparameters.}
Table~\ref{tab:hparam_approx} studies the approximation used on cache hits. OSI+Warp yields the strongest overall fidelity compared to OSI-only or warp-only, so we adopt it as the default OFA operator. For warping, a moderate scale (0.5) performs best, balancing alignment benefits with the noise sensitivity of high-resolution flow/warp signals.

Unless stated otherwise, these selected defaults are used for all main results.

\subsection{Effect of the number of denoising steps.}
We vary the denoising step budget from 35 to 140 and report end-to-end wall-clock latency (Fig.~\ref{fig:steps_ablation}). As expected, the unaccelerated baseline scales roughly linearly with the number of steps (57.0\,s at 35 steps, increasing to 199.1\,s at 140 steps), since each step requires a full DiT forward pass. WorldCache substantially reduces the effective per-step cost via cache reuse, yielding much lower latencies (25.0\,s, 34.2\,s, 45.5\,s, and 66.0\,s). The relative benefit is stable and becomes even more pronounced for longer trajectories, improving from \textbf{2.3$\times$} at 35 steps to a maximum of \textbf{3.10$\times$} among 70--140 steps. This indicates that caching opportunities grow with trajectory length, particularly in late refinement, where feature updates are small and the additional decision/approximation overhead remains minor.

\section{Additional Qualitative Results}
We provide additional Image2World examples on PAI-Bench that highlight the failure modes of naive cache reuse under dynamics and fine-grained interactions. In the driving scene (Fig.~\ref{fig:quality-1}), DiCache exhibits motion-related inconsistencies on salient moving pedestrians, including unstable appearance and slight ghosting near the end of the rollout, whereas WorldCache maintains more coherent identity and trajectories while preserving global scene layout. In the manipulation scene (Fig.~\ref{fig:quality-2}), the errors are even more localized and challenging as DiCache produces visible distortions around hands and the carried plate, having regions with fast, articulated motion and strong perceptual saliency, while WorldCache preserves object boundaries and hand poses consistency across frames. Together, these examples illustrate that WorldCache improves temporal coherence, particularly in the regimes most important for world models (foreground entities, contact-rich motion, and articulated interactions), aligning with our motion-aware decisions and improved cache-hit approximation.

\begin{figure}[t!]
    \centering
    \includegraphics[width=\textwidth]{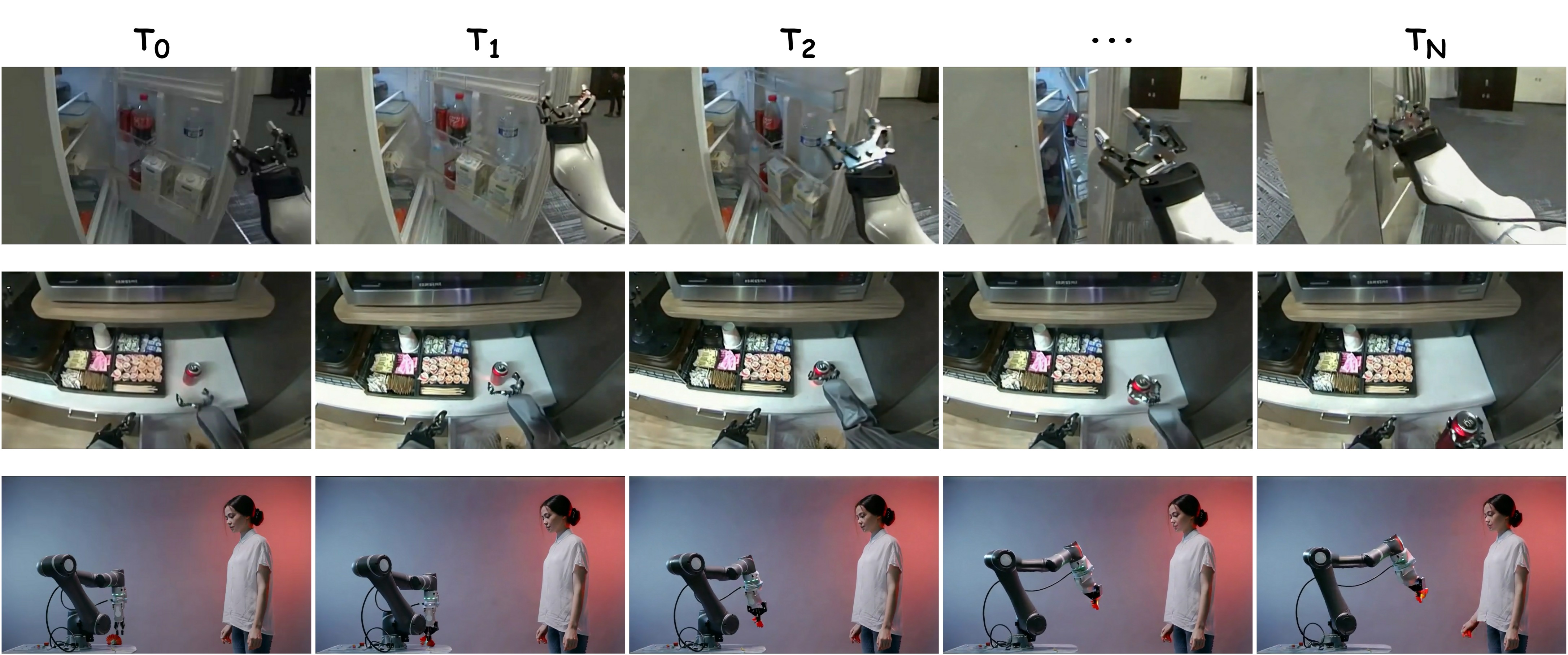}
    \caption{\textbf{Additional qualitative results.} We show representative frames at increasing rollout times $(T_0 \rightarrow T_N)$ from PAI-Bench (top to bottom). A robot hand interacting with a refrigerator door, an egocentric manipulation scene near a tabletop, and an arm–human interaction setup. Across all examples, the generated videos maintain stable scene layout and consistent object/robot geometry as motion evolves over time, demonstrating that WorldCache preserves temporal continuity in contact-rich and interaction-heavy rollouts. Better viewed with zoom-in.}
    \label{fig:quality-3}
\end{figure}

\begin{figure}[t!]
    \centering
    \includegraphics[width=\textwidth]{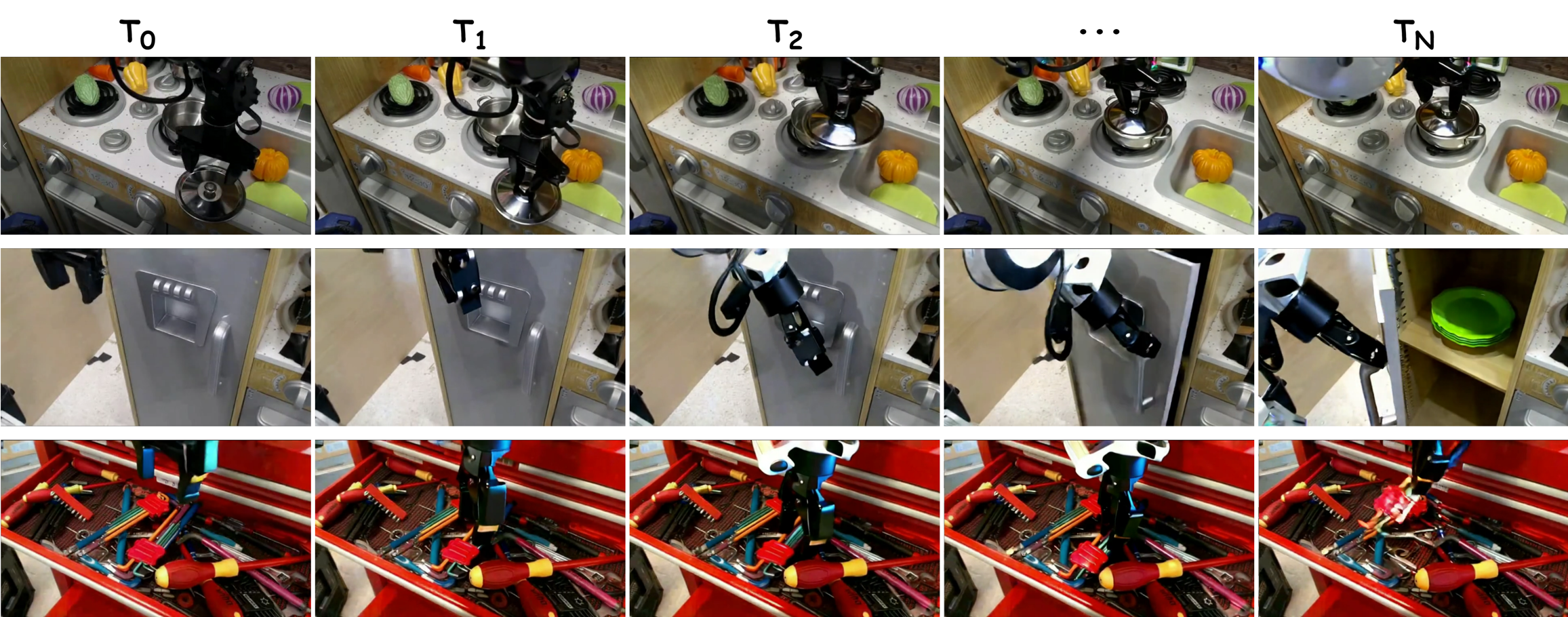}
    \caption{\textbf{Additional qualitative results.} We show evenly spaced frames from $(T_0\rightarrow T_N)$ for three representative scenarios: robotic arm usage for kitchen activities, opening a refrigerator door, and interaction within a cluttered toolbox. The sequences highlight sustained temporal coherence where robot pose, object geometry, and scene layout remain stable as the action unfolds. Better viewed zoom-in.}
    \label{fig:quality-4}
\end{figure}

Moreover, Fig.~\ref{fig:quality-3} and~\ref{fig:quality-4} provide additional robotics-centric I2V rollouts showing that WorldCache maintains stable scene structure and coherent robot or object motion over long horizons, including contact-rich interactions and cluttered environments. These examples complement PAI-Bench metrics by visually confirming that acceleration does not introduce drift in object geometry or kinematic consistency as the rollout advances.

\section{Limitations and Future Work}
\label{sec:limitations}

WorldCache is a training-free inference method. While our motion and saliency-aware constraints make caching conservative in difficult regimes, extremely abrupt scene changes (e.g., rapid viewpoint jumps, heavy occlusions) can occasionally reduce cache hit rates. 

A natural extension is to learn or adapt caching policies online, e.g., using lightweight predictors to estimate drift/saliency more accurately or to select per-layer/per-token reuse budgets. We also plan to integrate stronger motion estimation and uncertainty-aware warping to improve cache-hit approximation under high-speed dynamics and occlusions. 

\end{document}